\documentclass[manuscript]{acmart}
\usepackage{microtype}
\usepackage{graphicx}
\usepackage{subfigure}
\usepackage{booktabs} % for professional tables
\usepackage{tabularx}
\usepackage{multirow}  
\usepackage{multicol}

\usepackage{mathtools}
\usepackage{amsmath} % for maths like "cases"

\DeclareMathOperator*{\argmin}{arg\,min}
\usepackage{algorithm}
\usepackage{algorithmic}

\begin{document}

\title{A Deep Neural Networks ensemble workflow from hyperparameter search to inference leveraging GPU clusters}

\author{Pierrick Pochelu}
\email{pierrick.pochelu@totalenergies.com}
\affiliation{
  \institution{TotalEnergies SE}
  \city{Pau}
  \country{France}
}

\author{Serge G. Petiton}
\email{serge.petiton@univ-lille.fr}
\affiliation{
  \institution{Univ. Lille, CNRS, UMR 9189 CRIStAL}
  \city{Lille}
  \country{France}
}

\author{Bruno Conche}
\email{bruno.conche@totalenergies.com}
\affiliation{
  \institution{TotalEnergies SE}
  \city{Pau}
  \country{France}
}

\renewcommand{\shortauthors}{Pochelu et al.}





% HPC
\begin{abstract}
Automated Machine Learning with ensembling (or AutoML with ensembling) seeks to automatically build ensembles of Deep Neural Networks (DNNs) to achieve qualitative predictions. Ensemble of DNNs are well known to avoid over-fitting but they are memory and time consuming approaches. Therefore, an ideal AutoML would produce in one single run time different ensembles regarding accuracy and inference speed.
While previous works on AutoML focus to search for the best model to maximize its generalization ability, we rather propose a new AutoML to build a larger library of accurate and diverse individual models to then construct ensembles. First, our extensive benchmarks show asynchronous Hyperband is an efficient and robust way to build a large number of diverse models to combine them. Then, a new ensemble selection method based on a multi-objective greedy algorithm is proposed to generate accurate ensembles by controlling their computing cost. Finally, we propose a novel algorithm to optimize the inference of the DNNs ensemble in a GPU cluster based on allocation optimization. The produced AutoML with ensemble method shows robust results on two datasets using efficiently GPU clusters during both the training phase and the inference phase.
\end{abstract}

\maketitle

%HPC
Deep Neural networks (DNNs) are notoriously difficult to tune, train, and ensemble to achieve state-of-the-art results. Automatic machine learning with ensembling or "AutoML+ensembling" tools provide a simple interface to train and evaluate many ensembles of DNNs to achieve high accuracy by reducing overfitting. 
%It is a helpful tool for novices and machine learning practitioners to run those different tasks automatically: hyperparameter optimization and models selection.

% POPULARITY
Nowadays, multiple researchers and practitioners have well understood the benefit of ensembling DNNs. For example, in cyber-attack detection \cite{enscyb:2020}, time series classification \cite{enstim:2018}, medical image analysis \cite{ensimg:2021}, semi-supervision \cite{enssem:2021} and unbalanced text classification \cite{enstex:2020}.  Further, several winners and top performers on challenges routinely use ensembles to improve accuracy. However, ensembles of DNNs suffer from three main limitations to be widely deployed in research and industrial applications.

% How build the library?
% HPC
The first limitation is a lack of understanding about the best way to build base DNNs to construct an ensemble. Ensembling is still not fully understood \cite{kuncheva:2003} \cite{heteroinput:1999} \cite{creatediversity:2005} but authors generally agree that a large number of models, high diversity, and high individual performance are the three key components but we ignore in which proportion. That is why automatic procedures assessing multiple possible ensembles have been proposed. AutoML with posthoc ensembles \cite{autosklearn} works as follow: first, it automatically builds a library of hundreds of models, and then, it explores thousands of ensembles among the huge number of possible ensembles (combinations) \cite{caruana:2004} \cite{Caruana:2006} \cite{tsoumakas:2008} based on their validation score. However, it is still unclear how best to construct the best library of models to perform ensemble selection and only experimental evidence seems able to drive the algorithmic choices of the workflow. %Most of the previous AutoML researches works only on the accuracy optimization of non-deep machine learning. % In AutoML with posthoc ensembling we rather facing a different goal, we want to find the best library to combine models. 
%Even if the effect of diversity on ensemble accuracy is still not fully understood \cite{creatediversity:2005}, a major observation \cite{divermore:2008} is that parametric diversity is associated with hyper-parametric diversity enables to build more efficient ensembles.
% % IA
% The first limitation is a lack of understanding about the best way to build base DNNs to construct an ensemble. Multiple AutoML workflows have been invented aiming at vector classification or vector regression to build base models and ensemble them to reduce overfitting. They never have been generalized at scale on deep learning and yet those are the most complex models the most affected by overfitting due to a large number of parameters. We distinguish two categories according to how models are aggregated: post-hoc ensembling and ad-hoc ensembling. \\
% We distinguish also ad-hoc homogeneous ensembles consisting in construct $n$ different DNNs with the same hyper-parameters but at different run times. On the opposite, a  heterogeneous ensemble contains different DNNs build with different individual hyper-parameters, so the number of dimensions n of the ensemble hyperparameters space is multiplied compared to a  homogeneous ensemble. Homogeneous ensembles produce less diverse models (parametric diversity but the same hyperparameters) but reduce the ensemble hyperparameter space. Our work compares post-hoc ensembling with ad-hoc ensembling as a baseline.

% Computational cost
The second limitation is the lack of control of the computing cost of the produced ensemble.  In the previous works, authors apply ensembling \cite{caruana:2004} \cite{autosklearn} of non-deep machine learning and propose that the number of models to put in the ensemble as a threshold between accuracy and computing cost. In DNNs such as applied on image recognition, the combined models are heterogeneous with orders of magnitude of resource requirement. That is why the number of models as the constraint is not relevant to control the ensemble computational cost. Moreover, when generating automatically multiple models it is known that \cite{cloud:2017} there is no clear correlation between accuracy and DNNs cost, meaning that sometimes fast DNNs can be prioritized without significantly lowering the accuracy.

% HPC
Finally, no inference server enables the deployment of heterogenous DNNs and fully leverages modern GPU clusters. Current inference server allows to deploy deep neural networks \cite{tritonserv} \cite{ray} \cite{tfserv}, but the administrator of those servers has to attach manually DNNs to GPUs and set the batch size. Ensemble of DNNs is much more complex because we must deal with multiple DNNs sometimes co-localized into the same device and multiple devices. An ideal inference server of DNNs ensemble would allow computing automatically the best localization and batch size settings at the initialization phase.
% IA
% NOTHING

% HPC
It is time to address this missing piece in deep learning pipelines between AutoML, ensemble, and GPU clusters. To summarize our contributions, we run extended benchmarks with seven algorithms to generate the best library of models compared to 7 other algorithms and conclude that asynchronous Hyperband \cite{AHB:2020} suits this goal. After that the library is generated, we propose a new simple algorithm named SMOBF (scalarized-multi objective with budget greedy forward) to build an efficient ensemble based on their accuracy and a desired maximum computing cost. Third, we propose a novel server design to deploy with high efficiency, high flexibility, and low overhead a heterogeneous ensemble of DNNs.
% IA
%It is time to address this missing piece in deep learning pipelines between AutoML of Deep Neural Network and ensemble. To summarize our contributions, we run extended benchmarks with hyperparameter optimization, ensemble construction, and combination rule. We propose a new simple algorithm named SMOBF (scalarized-multi objective with budget greedy forward) to posthoc build an efficient ensemble based on their accuracy and a desired maximum computing cost.

This paper is as follows. In section II we go into further detail in the related fields: AutoML and AutoML+ensembling. In section (III) the different steps of the workflow AutoML+ensembling are analyzed and introduced. In section (IV) we introduce the inference server design for deployment. In section (V) we analyze our AutoML procedure then (VI) we benchmark our inference server on 2 generated ensembles.

\section{Analysis of the AutoML field}
\label{sec:rel}

\subsection{AutoML}

% AutoML
The empirical nature of research in Deep Learning leads scientists to try many model architecture settings, optimization settings, and pre-processing settings to find the best-suited one for data. AutoML is made of 3 modules, the DNNs search space (the ``hyperparameter space''), the DNNs sampling strategy (or ``hyperparameter optimization''), and the evaluation phase consisting in returning the score associated with one hyperparameter value.

% Definition formelle
Any model previously trained with hyperparameter $\lambda$ sampled from hyperparameter space $\Lambda$ is written $M_{\lambda}$. The hyperparameter optimization goal defined in equation~\ref{eq:hpo} consists in finding the best hyperparamer $\lambda*\in\Lambda$ building a model to reduce the error measured by $E$. The error is measured on the validation data $x_{valid}$ matching labels $y_{valid}$. 
\begin{equation}
    \lambda * = \argmin_{\lambda\in\Lambda} E(M_{\lambda}(x_{valid}),y_{valid}) 
    \label{eq:hpo}
\end{equation}

% Difficulty of the field
In the literature, AutoML algorithms are typically compared based on their results in the evaluation phase. While this may seem intuitive, this field is now facing multiple methodological questions \cite{random:2020} on the relevance of comparing multiple workflows made of different stages and different initial conditions. And more, to fairly compare their robustness, multiple datasets, and multiple random seeds must be reported which is a computing-intensive research area limiting initiatives.

Our paper does not claim the superiority of our AutoML+ensembling method in all cases but it is rather the first step toward AutoML+ensembling of DNNs aiming at a desired trade-off between the produced ensemble accuracy and its cost for practical usage. Our work is applied on the computed vision task but can be generalized on other tasks where qualitative prediction is required leveraging one or more clusters of GPUs.

%Plenty of methods
No Free Lunch theorem \cite{NFL:1997} proves that no black-box hyperparameter optimization (HPO) can show superior performance to random search in all cases. Nevertheless, methods searching between global exploration and local exploitation have shown a stable performance on diverse applications. Early stopping \cite{early:1998} \cite{HB:2017} has been also proposed to stop the less promising trials to focus hardware on the most promising trials accelerating the overall optimization process. 

%ods capable to evaluate different trials at the same time and increase the overall convergence. Those methods are especially effective on the most exploratory methods to evaluate large populations at the same time.

%Furthermore, many parallel methods have been proposed enabling assessing multiple trials at the same time and leveraging the. Method exploring multiple trials and Highly scalable and optimization Robust method is the desired features. Hardware efficient methods are expected to find better optimum than method wasting time.
% Parallel random search on $n$ GPUs should found the optimum faster than
% Parallelism is a major 
% Parallel method capable to assess different trials at the same time is also a desired algorithmic property. n Random searches at the same time should found the best value n times faster compared to the sequential version.

\subsection{AutoML frameworks}

% Sequential
Recently, AutoML methods sequentially updating the current DNN (neural network morphism) have been proposed based on Bayesian optimization (Auto-Keras \cite{autokeras}), reinforcement learning (ENAS \cite{ENAS:2018}) or differentiable architectures (DARTS \cite{darts:2018}). These methods require several hours on a single modern GPU and converge to a sub-optimal solution.

% HPO+ parallel SMBO
A more general approach consists in searching not only neural network architectures but optimization settings and data processing too with fixed-length vector hyperparameters. Bayesian optimization, like Sequential Model-Based Optimization (SMBO) with Gaussian model is known to perform well to optimize continuous hyper-parameters \cite{snoek:2012} \cite{bogp:2011}. Tree-based models are more adapted to the discrete hyperparameters like Tree Parzen Estimator (TPE) \cite{TPE:2011} and Sequential Model-based Algorithm Configuration (SMAC) \cite{SMAC}. Despite that SMBO are inherently sequential methods, a parallel version have been proposed \cite{snoek:2012} based on successive populations of trials to explore the hyperparameter space by leveraging multi-cores.

% GENETIC
Evolutionary methods \cite{evolution:2017} \cite{evolution:2020} are naturally parallelizable algorithms running successive populations (called ``generation'') of trials in parallel. By running them on a large scale on GPU clusters for several days, authors discovered those methods can converge very late and very high with a high robustness.

\subsection{AutoML with ensembling}

AutoML which not only selects the best model but combines them is valuable for domains that require the best possible accuracy. Several benchmarks were performed and AutoML+ensembling is today considered as the big AutoML challenge series winner on data points like AutoML challenges \cite{Guyon2019} and Kaggle challenges on image recognition. Previous researches on non-deep machine learning ensembles shows that over-fitted machine learning algorithms predictions can be averaged out to get more accurate results \cite{useoverfit:1995}. This phenomenon is mainly explained by the Law of Large Numbers which claims that the average of the results obtained from many non-biased trials should be close to the expected value. These results are especially interesting for deep learning models. They are the most affected models to random effects (over-fitting) due to their huge number of parameters.

% Ad-hoc versus posthoc ensembling
We observe two main trends in AutoML with ensembling. First, \textit{AutoML+Ad-hoc ensembling} \cite{autoweka} \cite{h2oautoml} \cite{fleet} \cite{autostacker} which consists in searching directly ensembles. The hyperparameter space describes an ensemble of fixed size. At the end, the HPO algorithm keeps the best ensemble and wastes all the others. It suffers from a lack of flexibility because the number of models in an ensemble is fixed before build all DNNs. In other words, changing the number of DNNs in the ensembles require a new AutoML runtime. Second, \textit{AutoML+Post-hoc ensembling} \cite{caruana:2004} \cite{tsoumakas:2008} \cite{autosklearn} runs a standard HPO algorithm providing a library of trained models, then constructs ensembles from the library based on a greedy algorithm. This approach is flexible because we can produce several ensembles with different numbers of models with the same library of DNNs. 

% POUR AAAI
%The posthoc ensembling allows computing more ensembles because each DNNs model can be used for multiple ensembles. The ad-hoc models are computed with the formula $\lceil \frac{L}{S} \rceil$ with the desired size of ensembles $S$ and $L$ the library size. In comparison, the number of possible combinations without replacement of posthoc ensembling is the known combination formula given by $\frac{L!}{S! * (L - S)!}$. Therefore, when using an exploratory-only HPO such as Random Search there is no benefit to using ad-hoc ensembling. In exploratory-exploitative such Bayesian framework exploring in the ensemble hyperparameter space,  this claim is not possible because the optimization is done in the ensemble hyperparameter space and the previous iteration allows to compute the next ensemble construction.

\section{Proposed workflow}
\label{sec:work1}

\begin{figure*}[]
    \centering
    \includegraphics[width=\linewidth]{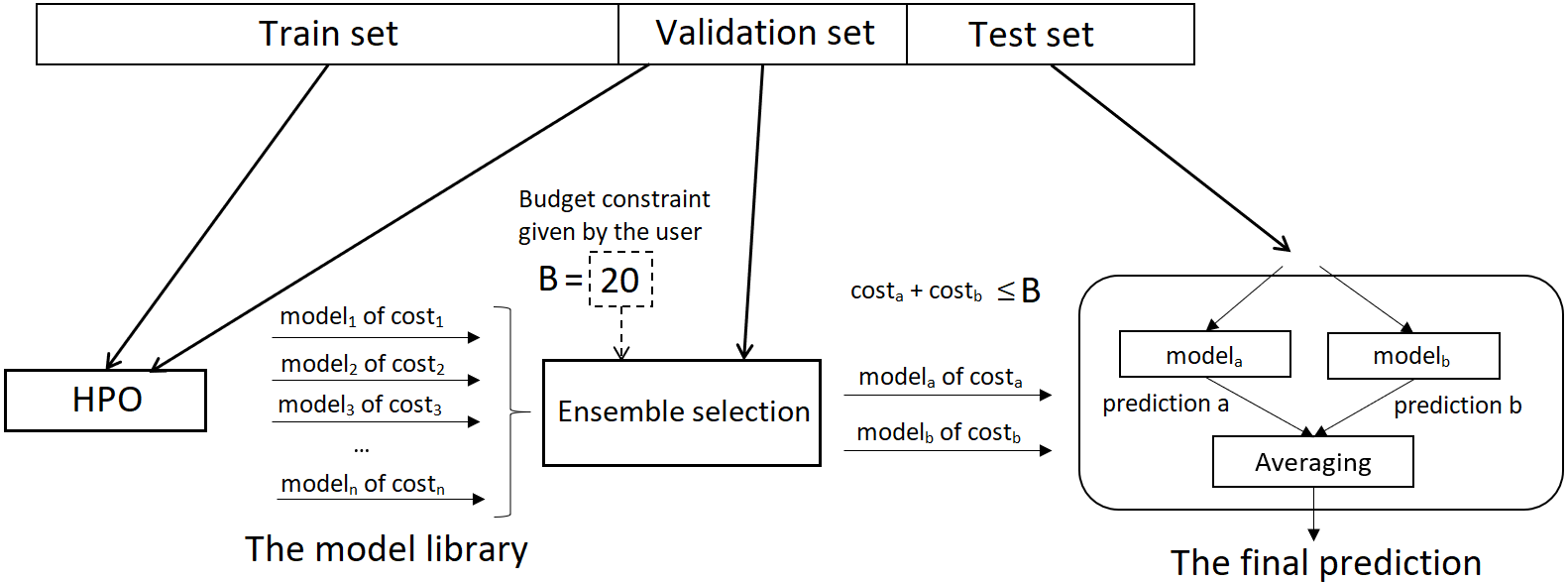}
    \vspace{-0.7cm}
  %HPC
  \caption{The proposed workflow runs 3 steps 1) The HPO algorithm generates a library of models. The trials are distributed on several GPUs synchronized by a master process. We recommend asynchronous Hyperband based on experimental results. 2) We propose a new multi-objective Ensemble Selection algorithm to search the most efficient ensemble honoring a budget given by the user (here B=20). It is based on a parallel greedy algorithm evaluating hundreds of combinations per second on multi-core CPUs. 3) The returned ensemble predicts by combining (averaging) DNNs predictions on new data. The deployment of the ensemble into a server is not shown in the figure.}
%   % IA
%   \caption{The proposed workflow runs 3 steps 1) The HPO algorithm generates a library of models. The trials are distributed on several GPUs synchronized by a master process. We recommend asynchronous Hyperband based on experimental results. 2) We propose a new multi-objective Ensemble Selection algorithm to search the most efficient ensemble honoring a budget given by the user (here B=20). 3) The returned ensemble predicts by combining (averaging) DNNs predictions on new data.}
 \label{fig:workflow}
\end{figure*}

% HPC
In this section, we give an overview of the proposed workflow~\ref{fig:workflow}. Then, we will go into further details on the distributed architecture that accelerates these three steps of our AutoML workflow: HPO to construct a library of models, Ensemble Selection to construct ensembles based on their validation accuracy and computing cost, and ensemble deployment.
% IA
% In this section, we give an overview of the proposed workflow~\ref{fig:workflow} before understanding the nature of our work. Then, we will go into further details in each step of the proposed AutoML workflow: HPO to construct a library of models, Ensemble Selection to construct ensembles based on their validation accuracy and computing cost, and finally the combination rule.

\subsection{Detail of the workflow}

% Objective
\textbf{Hyperparameter optimization} (HPO). The choice of the HPO algorithm has several impacts in terms of the number of produced models at a given time horizon, the accuracy of models, and their diversity. After several experiments, we recommend Hyperband based on our experiments.

% HPC
In this regard, Hyperband produces the biggest library of models among tested algorithms because most models do not reach the maximum number of epochs. It is also a lock-free distributed algorithm until the termination of the algorithm, spending most of its runtime to train models occupying all GPUs and storing them on the library. Then, Hyperband produced also very diverse models due to the initial random sampling of hyper-parameters which is desirable in terms of final accuracy. Finally, because more training iterations are given to the best models, Hyperband performs explore-exploit in the time (not in the hyperparameter space) and we observe it builds a better distribution of individual DNNs (regarding the top 10 \% and top 25 \%) compared to random search on all our runtimes. But we cannot expect that Hyperband outperforms or underperforms other exploratory-exploitative HPOs in all cases such as Bayesian or genetic algorithms \cite{NFL:1997}. A large number of models, hyperparametric diversity, and efficiency are three reasons explaining why Hyberband is robust to build a library of models but we still ignore in which proportion they are important.
% IA
% In this regard, asynchronous Hyperband \cite{AHB:2020} produces the biggest library of models among tested algorithms because most models do not reach the maximum number of epochs. It is also a lock-free distributed algorithm until the termination of the algorithm, spending most of its runtime to train models occupying all GPUs and storing them on the library. More models intuitively increase the probability to found good combinations between them during the ensemble selection phase. Then, Hyperband produced also diverse models hyperparameters due to the initial random sampling of hyper-parameters. Finally, Hyperband performs explore-exploit in the time (not in the hyperparameter space) but we cannot expect that Hyperband outperforms or underperforms other HPOs in all cases \cite{NFL:1997}. A larger number of models, hardware occupancy, and hyperparametric diversity are three reasons explaining why Hyberband is robust to build a library of models but we still ignore in which proportion they are important. Detailed analysis of the produced ensembles in terms of bias-variance-covariance decomposition \cite{creatediversity:2005} and distribution would allow providing more insight.

\textbf{Ensemble selection.} An ensemble selection algorithm finds the best combination possible honoring the budget given by the user. We propose SMOBF greedy standing for "Scalarized Multi-Objective with Budget Forward greedy".

%Forward greedy
Forward greedy \cite{caruana:2004} starts with the empty ensemble and successively adds the best available model to improve the ensemble target metric. The algorithm stops when no available model can improve accuracy or respect the budget.

% SMOBF
We propose new equation~\ref{eq:smo} to inform the greedy algorithm to favor accurate and cheap DNNs before consuming the overall budget. The current ensemble is $a$ with its computing cost $C_{a}$ and its predictions $y_{a}$. Penalty $P_{a}$ returns 0 when the budget $B$ is honored ($C_{a} \leq B$ ) or an arbitrary large number when it is not to exclude $a$ from the possible ensembles.

The weight $w$ allows controlling the nature of the solution found by the greedy algorithm by placing greater or lesser emphasis on the objectives. The greedy algorithm is run multiple times with different values $w=0.1$, $w=0.01$, and $w=0.001$. Scalarization is a convenient way to handle multi-objective problems by reducing them to a single objective problem, so a simple mono-objective optimization can be performed.
\begin{equation}
    score_{a} = (1-w) * E( y_{a} , y ) + w * C_{a} + P_{a} 
    \label{eq:smo}
\end{equation}

% SMOBF
Then, the best ensemble is picked according to its validation cross-entropy loss and respecting the given budget. To improve the robustness of the method on new applications $C\_{a}$ is standardized and it is the rate between the sum of computing cost of base models (time to predict 2000 images) and the budget $B$.

%Cross validation
Ensemble selection algorithm computes the validation loss of candidate ensembles to evaluate how well a solution will generalize on the test database. Since the data used for validating is taken away from training the individual models, keeping the validating set small is important. Smaller validating sets are however easier to overfit. Contrary to common AutoML on data points datasets, due to the cost of training and evaluating one model on images datasets, we do not repeat the experience with K-fold cross-validation.

% Pruning + Caching
In the library of models, some models diverge or have such poor performance compared to other models that they are unlikely to be useful to improve any ensemble building \cite{Caruana:2006}. Eliminating these models from the library should not reduce the performance and facilitates the ensemble selection task by decreasing the number of non-promising combinations. Pruning works as follows: models are sorted by their performance on the validation metric and only the top X\% of them are used for ensemble selection. After pruning, the predictions on the validation set are cached before running the ensemble selection algorithm. This allows handling only predictions vector and not models during the ensemble selection process.

\textbf{Ensemble combiner rule.} We use the simple average as a combiner rule. More advanced methods exist such as "ensemble selection with replacement" \cite{caruana:2004}, weighted averaging, and stacking. Those methods are calibrated on a validation set and thus prone to over-fitting.

Now that the workflow is developed, the final accuracy depends on two settings. The tuning effort of the library of models and the budget given by the user to generate a new ensemble.

\subsection{Distributed HPO with GPU clusters}

To assess large numbers of hyperparameter trials in a reasonable amount of time, a distributed framework to use one GPU cluster or several GPU clusters is required. Several trials are distributed with a middleware containing one scheduler and multiple workers, each worker being associated with heavy computing resources. Typically, the scheduler sends a hyperparameter set to the available workers with the number of training epochs to perform. Then, it gathers the scores of the trials from the workers and finally computes the next hyperparameters to send and so on.

% S'envoyer des fleurs
This distributed framework suits the need of most optimization by leveraging GPUs and clusters. Each trial is run independently without slowdown among themselves. Workers can connect or disconnect to the master at run time; however, a disconnection interrupts the current trial. Many HPO algorithms benefit from this middleware such as random search, Hyperband, parallel SMBO, evolutionary algorithms, and also some greedy algorithms for discrete optimization such as SMOBF.

% 1 trial
Modern clusters have evolved into a hybrid machine that contains both CPUs and GPUs on each node. These heterogeneous CPU-GPU clusters are particularly useful to accelerate the training loop of one trial. One trial is implemented with two processes, the \textit{preprocessor} which loads and preprocesses the data on multi-cores, while the \textit{trainer} trains the DNNs on the GPU. In case the entire training dataset fits into memory, the preprocessor does not need to load multiple time the data, it is shared between preprocessors of the cluster to avoid copies and useless I/O.

\subsection{Distributed Ensemble Selection}

Ensemble selection is a greedy algorithm evaluating all neighbors of a current ensemble at each iteration.  For each ensemble $a$, the equation \ref{eq:smo} is performed. This procedure takes one second on one CPU with 100 elements per prediction (100 classes), 2000 validation data samples, and a relaxed budget.

In the case of semantic segmentation images, the class predictions are at pixel scale making this procedure much more computing-intensive. A typical dataset for autonomous vehicle applications \cite{citydb:2016} contains 256x256 input images, 30 classes per pixel and 500 validation samples. Linear scaling indicates that the computing cost of this procedure would take 1h20 (256x256x30x500 predictions elements compared to 100x2000 predictions elements in CIFAR100 taking 1 second). Because SMOBF is an optimization algorithm we may use a similar distributed framework (previous section) to distribute neighbors evaluation on CPUs or GPUs to alleviate this computing cost.

\subsection{Multi-GPU inference}
\label{sec:alloc}

\begin{figure}
        \centering
        \includegraphics[width=0.5\linewidth]{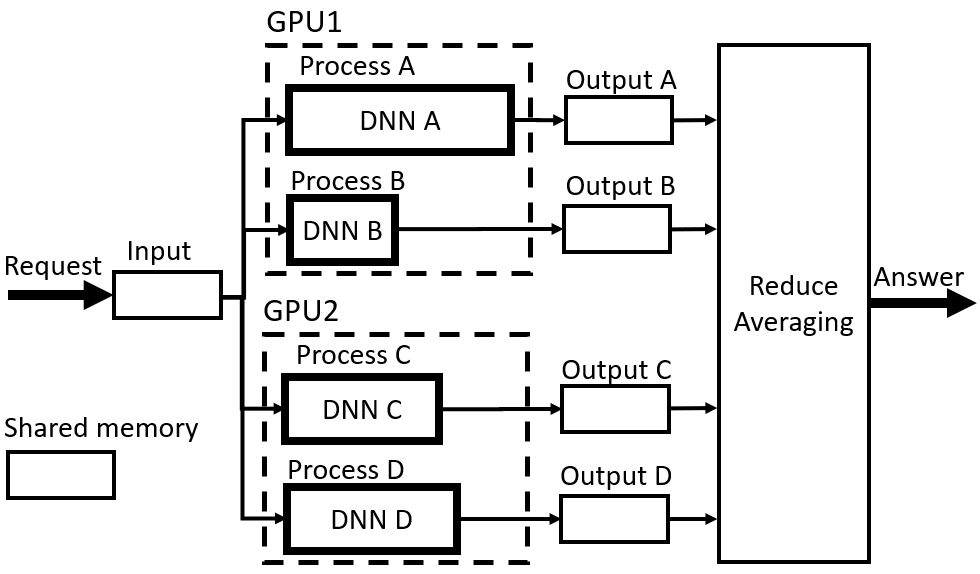}
        \caption{Toy example of allocation of 4 heterogenous DNNs into 2 GPUs for inference. There are 5 processes, the orchestrator containing and its 4 children A, B, C and D. The orchestrator receives data to predict and serve predictions as a service. Shared memory are buffers in the RAM.}
        \label{fig:alloc}
\end{figure}

% Inference pipeline
After getting an ensemble of DNNs, we need to serve it efficiently on the available computing resources. We design an efficient inference pipeline illustrated in figure \ref{fig:alloc}. All DNNs into the ensemble predict asynchronously. We store the input and the outputs into shared data to avoid slowing copies. The orchestrator asynchronously runs in one CPU-core the cumulative averaging of all predictions to avoid slowing down the entire pipeline.

Algorithm 1 aims to return an available solution in terms of memory while algorithm~\ref{algo:speed} speeds up the allocation settings found based on iterative updates.

%We demonstrate in the previous section that we cannot evaluate all possibilities to allocate an ensemble of $n$ DNNs, $B$ batch size values and $M$ devices settings which are counted with the formula with of that is why we propose an approximate solution made of two algorithms. 

% Presenter l'algo1
Algorithm 1 balances the memory requirement between devices to fit into memory in only $n$ steps, with $n$ the number of models in the ensemble.  It is basically a variant of the worst-fit-decreasing algorithm to solve a bin packing problem except we give a priority on the GPU because we make the assumption it is faster. If it is not possible, it allocates on the CPU because more memory is present. We decided to not write it for space reasons and because it is similar to the already known worst-fit-decreasing.

% Presenter l'algo 2
Algorithm~\ref{algo:speed} starts with a correct allocation and at each iteration, it assesses all neighborhood settings distant from 1 update. One update consists of either updating the batch of any DNN or changing its device. The algorithm is stopped when no neighborhood improves the target metric or when the max number of iterations is reached. The number of possible combinations is $(M+B)^{n}$  with $M$ devices, $B$ batch size values, and $n$ DNNs in the ensemble. Algorithm~\ref{algo:speed} breaks down this complexity into a succession of $Mn+Bn$ combinations (or ``neighbors'') to assess. We have no guarantee that this greedy algorithm finds the optimal allocation and we cannot verify if it is obtained (except if we brute-force all possibilities). However, we have the guarantee that in the worst-case scenario, a solution as good as the starting one is returned (line 10). 

% Expliquer le formalisme de struct
We formulate an inference setting like a $\{B,G,C\}$ set containing three lists. We use ordered structures because it makes it easier to write the algorithm. $B_{i}$ is the batch size of the $i^{th}$ DNN. $G_{j}$ is the set of DNNs contained in the $j^{th}$ GPU, in the same way, $C_{k}$ contains DNNs into the $k^{th}$ CPU. Again, a DNN is placed into one single device but a device can contain multiple DNNs.

% Expliquer les functions
 In algorithm~\ref{algo:speed} $I$ are randomly generated images to benchmark and calibrate the inference server. The number of fake images must be chosen high enough to smooth random effect and not too high to reduce the duration of the overall algorithms. $bench$ function instantiates the pipeline with the given allocation settings (first argument) on the fake images (second argument) and returns the performance metrics (e.g. the number of predictions per second). The same algorithm can generalize well to other performance metrics such as latency consisting in reducing the time between one input data sample and its prediction.

\begin{algorithm}[]

\caption{Refine GPUs allocation to speed up}
\label{algo:speed}
\begin{algorithmic}[1]

\STATE{\textbf{input:} $D$ the list of DNNs in the ensemble, $PB$ possible batch size values, $max\_combi$ maximum number of assessed combinations, $G$ and $C$ are preliminary GPU and CPU allocation, $B$ preliminary batch sizes}
\STATE{\textbf{output:} $G$, $C$, $B$}
\STATE{\textbf{start}}

\STATE{ $I$ $\gets$ $fake\_images()$ // generate fake data to calibrate the allocation }

\STATE{ $current\_score$ $\gets$ bench($D$,$B$,$G$,$C$,$I$)}

\WHILE{$trials < max\_combi$}

\STATE{better\_allocs $\gets$ []}
\STATE{better\_scores $\gets$ []}

\STATE{$i\_allocs$  $\gets$ update\_i\_alloc($i$,$B$,$G$,$C$) }

\FORALL{$\{B2, G2, C2\}$ in $i\_allocs$}
\IF{$\{B, G, C\} \neq \{B2, G2, C2\}$}
\STATE{$score2$ $\gets$ bench($D$,$B2$,$G2$,$C2$,$I$)}
\IF{$score2 > current\_score$}
    \STATE{$better\_scores$.append($score2$)}
    \STATE{$better\_allocs$.append($\{B2, G2, C2\}$)}
\ENDIF
\IF{ $trials \geq max\_iter$ }
	\STATE{break}
\ENDIF

\ENDIF
\STATE{$trials \gets trials + 1$}

\ENDFOR

\IF{ length($better\_allocs$) > 0 }
    \STATE{$id$=argmax($better\_scores$)}
    \STATE{$\{B, G, C\}$ = $better\_scores$[$idbest$]}
    \STATE{$current\_score$=$better\_scores$[$idbest$]}
\ENDIF

\ENDWHILE

\STATE{return  $\{B, G, C\}$}

\end{algorithmic}

\end{algorithm}

\section{Experiments and results}
\label{sec:exp}

We experiment and discuss our workflow by varying the three steps of the AutoML workflow of deep neural networks on CIFAR100 and the microfossils datasets. More details is given in appendix~\ref{sec:appdesign}.

%It is possible that larger hyperparameters value ranges may positively influence again the results obtained.

%In this section, we evaluate different workflows by evaluating various HPO strategies, different combinatorial optimizations. We also assess different settings like the budget influence and tuning time influence.

\subsection{The infrastructure}

%HPC
Experiments were done on IBM Power9 architecture, containing 2 sockets. In each socket, there are 18 cores of CPU with a maximum frequency of 3.8Ghz and 256G of RAM. There are also 6 GPUs by node and are Nvidia Tesla V100-SXM2 with 16G of memory. Hyperparameter optimization framework Tune~\cite{tune:2018} was used. It runs above the Ray framework \cite{ray}, it schedules and spreads experiments to run on GPUs and store results into files. Deep Learning training loop and data augmentation was coded with the framework Keras \cite{keras:2017} with TensorFlow 1.14.0 \cite{tf:2020} backend. 
%IA
%Experiments setting are presented in appendix~\ref{sec:appdesign}.

\subsection{step 1 - HPO}
\label{sec:hpores}

% HPC
\subsubsection{Comparison of hardware allocation}
Our CNN framework based on ResNet can take between 15 seconds and 11min30 regarding the complexity of the neural architecture assessed (number of filters per convolution, number of convolutional blocks, ...). 
\begin{table*}[]
\setlength{\tabcolsep}{2pt}
\begin{tabularx}{\linewidth}{l|lllll|lllll}
\toprule
                & \multicolumn{5}{c}{9 trials 100 epochs}                                     & \multicolumn{5}{c}{30   trials 30 epochs}                                   \\
                & acc (\%)                        & duration & speed up & \#epochs & gpu (\%) & acc (\%)                        & duration & speed up & \#epochs & gpu (\%) \\
\midrule
RS 6GPUs        & $67.08\pm3.85$                  & 10h13    & 6.2      & 900      & 85       & 64.90+-0.92                     & 11h40    & 5.5      & 900      & 82       \\
\hline
HB 6GPUs        & \multirow{2}{*}{$62.58\pm4.73$} & 5h48     & 4.4      & 710      & 49       & \multirow{2}{*}{$62.94\pm5.28$} & 1h40     & 5.2      & 388      & 62       \\
HB 4nodes*6GPUs &                                 & 2h00     & 8.8      & 710      & 33       &                                 & 0h48     & 10.9     & 388      & 34       \\
\hline
GA 6GPUs        & $61.85\pm2.55$                  & 14h29    & 2.8      & 900      & 27       & $65.33\pm2.33$                  & 10h01    & 4.1      & 900      & 57      \\
\bottomrule
\end{tabularx}
\caption{Limited scale benchmarks of some HPO algorithms. Hyperband was set up with a halving of 3 and the genetic number of trials is divided into 3 successive generations. The columns from the left to the right are: The accuracy error which is not dependant on the number of resources, the duration of the HPO process, the speed up measured compared to the single GPU version, the  \#epochs is the number of train iteration performed, the mean percentage of occupied GPU during all the entire HPO process.}
\label{tab:bench}
\end{table*}
Our benchmark \ref{tab:bench} reveal that Random Search and Hyperband occupy more the GPUs than algorithms based on a sequence of population algorithms (such as GA) which is explained by the fact that no intermediate stopping of all GPUs is required. Those benchmarks also confirm that Hyperband terminates earlier than Random Search because the less promising DNNs have been stopped before the maximum number of epochs is reached. The scalability of HPO algorithms and saving useless computations are two important algorithmic characteristics to explore a large number quantity of DNNs. \\
At the end of any HPO algorithm, the last few DNNs free the GPUs but they take various amounts of time to terminate, this explains why we do measure not a perfect usage of GPUs of Random Search and Hyperband. Hyperband emphasis this phenomenon because the early stopping increases the variability of time took between DNNs.

\subsubsection{HPO to produce ensembles}

We compare in table~\ref{tab:comphpofos} our presented workflow by varying the HPO algorithm to generate a different library of models. Random Search (RS) \cite{rs:2012}, asynchronous Hyperband (HB) \cite{AHB:2020}, Bayesian Optimization with Gaussian Processes (BOGP) \cite{bogp:2011}, BOHB, Sequential Model-based Algorithm Configuration (SMAC) \cite{SMAC}, Tree Parzen Estimator (TPE) \cite{TPE:2011}, Genetic Algorithm (GA). To combine generated models, we benchmark them with the SMOBF greedy ensemble selection algorithm. We benchmark three times each algorithm and we show the median value of each computing. When the budget is relaxed we observe the standard deviation is generally lower than 0.1\%. When the budget is lower the standard deviation is lower than 1.5\%.

\begin{table}[]
\centering
\small
\begin{tabularx}{10cm}{ll|ccccccc}
\toprule
Data & Budget    & RS             & HB             & BOGP  & BOHB           & SMAC           & TPE   & GA             \\
\midrule
\multirow{8}{*}{C100} & 20  & 31.26          & 27.8           & 31.58 & 29.24          & \textbf{24.44} & 29.97 & 24.61          \\
~ & 40  & 24.24          & \textbf{22.87}          & 28.2 & 25.6           & 22.97 & 26.01 & 22.91          \\
~ & 60  & 22.27          & 21.85          & 26.91 & 23.53          & 22.65          & 25.44 & \textbf{21.51} \\
~ & 80  & 22.69          & 21.73          & 26.2  & 22.58          & 22.17          & 24.25 &  \textbf{21.07} \\
~ & 120 & 21.63          & \textbf{21.55}          & 24.78 & 21.86 & 22.17          & 23.03 & 21.95          \\
~ & 160 & \textbf{20.11} & \textbf{21.11} & 24.24 & 21.64          & 21.87          & 22.47 & 21.91          \\
~ & 240 & 20.7           & \textbf{20.46}          & 23.76 & 21.2           & 21.87          & 22.55 & 21.91          \\
~ & 320 & 20.64          & \textbf{20.42} & 23.76 & 21.1           & 21.87          & 22.46 & 21.91         \\
\hline
\multirow{6}{*}{Micro} & 125  & 13.45 & 11.93         & 14.44 & 12.18 & \textbf{10.27} ~ & 13.91 & 10.33 \\
~ & 250  & 10.98 & 9.82          & 10.93 & 11.71 & \textbf{9.63}  & 10.66 & 9.71  \\
~ & 375  & 10.84 & \textbf{9.32} & 9.93  & 10.5  & 9.45           & 10.49 & 9.5   \\
~ & 500  & 10.23 & \textbf{8.91} & 9.93  & 9.84  & 9.27           & 9.06  & 9.28  \\
~ & 750 & 9.7   & \textbf{8.87} & 9.21  & 9.28  & 9.18           & 9.3   & 9.18  \\
~ & 1000 & 9.69  & \textbf{8.46} & 8.77  & 8.8   & 9.09           & 9.21  & 9.07 \\
\bottomrule
\end{tabularx}
\caption{Our workflow error (\%) by comparing seven HPO algorithms was run on both datasets for 6 days on 6 GPUs. Each HPO generates a library of models for a given dataset. The budget is expressed as "sum of DNNS times (seconds) to predict 2K images on 1 GPU"}
\label{tab:comphpofos}
\end{table}

Not surprisingly, when the budget increases the ensemble selection can find better ensembles from any library. This is explained by the fact that the number of available good combinations between them increases. In the table~\ref{tab:comphpofos} some algorithms converge and do not found better ensembles after high budgets such as the Bayesian method and genetics ones. Different run time shows similar conclusion.  On both datasets, the best ensemble is found with Hyperband. 79.44\% accurate on CIFAR100 and 91.54\% accurate on microfossils. We do not show results of AutoML without ensembling compared to AutoML+ensembling due to a lack of space, but AutoML+ensembling is Pareto dominant for all budgets and all performed run times.

\subsubsection{Effect of tuning time}
\label{sec:tunetime}

\begin{figure}[]
    \centering
    \begin{minipage}[]{0.48\linewidth}
        \centering
        \includegraphics[width=\linewidth]{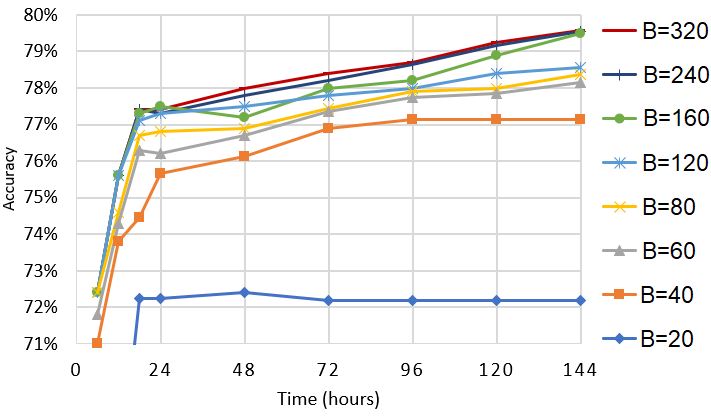}
        \vspace{-1cm}        
        \caption{Ensemble from Hyperband algorithm on the CIFAR100 dataset}
        \label{fig:automltime1}
    \end{minipage}
    %\vfill
    \begin{minipage}[]{0.48\linewidth}
        \centering
        \includegraphics[width=\linewidth]{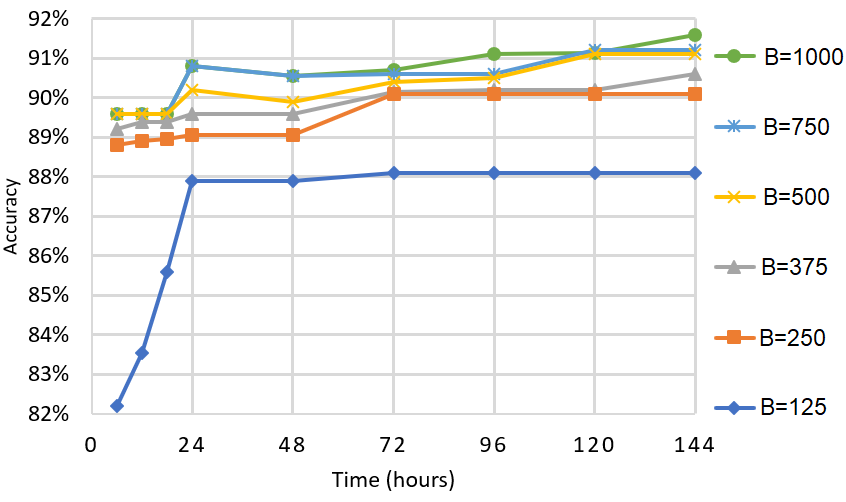}
        \vspace{-1cm}
        \caption{Ensemble from Hyperband on the microfossils dataset}
        \label{fig:automltime3}
    \end{minipage}
    % \vfill
    % \begin{minipage}[t]{\linewidth}
    %     \centering
    %     \includegraphics[width=1.0\linewidth]{img/automltime3.PNG}
    %     \vspace{-0.8cm}
    %     \caption{Ensemble from Hyperband on the microfossils dataset}
    %     \label{fig:automltime3}
    % \end{minipage}
    % \vfill
    % \begin{minipage}[t]{\linewidth}
    %     \centering
    %     \includegraphics[width=1.0\linewidth]{img/automltime4.PNG}
    %     \vspace{-0.8cm}        
    %     \caption{Ensemble from SMAC algorithm on the microfossils dataset}
    %     \label{fig:automltime4}
    % \end{minipage}
\end{figure}

% presenter
In AutoML we often want to see the evolution of performance over the tuning time, not only the final performance after a time horizon. Therefore, we assess the workflow varying the budgets until it does not affect the ensemble construction anymore and at different tuning time snapshots. Those figures are presented in figure~\ref{fig:automltime1} \ref{fig:automltime3}.

% Time versus accuracy
First, regarding the benefit of the tuning time, we observe two main trends. When the budget is very small (B=20) models accuracy converges early to 18, 24, or 48 hours reaching the limit of the hyperparameter optimization with a little (or without) ensembling. When the budget is bigger the exploding number of available combinations of ensembles leads to the discovery of better ensembles. The post-hoc ensembling is a promising line of research that deserves more attention and more understanding of how DNNs interact with each other.
 
% More budget = more accuracy
Then, we observe that increasing the budget systematically leads to an increasing accuracy (colored lines are rarely crossed) but this trend decline. For example, in the figure~\ref{fig:automltime1} we show that when Hyperband is finished, the benefit is obvious from 1 to 2 models: +5 point of accuracy percentage, but the improvement is small from B=120 to B=320: +1 point of accuracy percentage.

% \begin{figure}[]
%     \centering
%     \begin{minipage}[t]{.485\linewidth}
%         \centering
%         \includegraphics[width=\linewidth]{img/automltime1.PNG}
%         \vspace{-0.8cm}        
%         \caption{Ensemble from Hyperband algorithm on the CIFAR100 dataset}
%         \label{fig:automltime1}
%     \end{minipage}
%     \hspace*{\fill}
%     \begin{minipage}[t]{.485\linewidth}
%         \centering
%         \includegraphics[width=\linewidth]{img/automltime2.PNG}
%         \vspace{-0.8cm}
%         \caption{Ensemble from SMAC on the CIFAR100 dataset}
%         \label{fig:automltime2}
%     \end{minipage}
%     \hfill
%     \begin{minipage}[t]{.485\linewidth}
%         \centering
%         \includegraphics[width=\linewidth]{img/automltime3.PNG}
%         \vspace{-0.8cm}
%         \caption{Ensemble from Hyperband on the microfossils dataset}
%         \label{fig:automltime3}
%     \end{minipage}
%     \hspace*{\fill}
%     \begin{minipage}[t]{0.485\linewidth}
%         \centering
%         \includegraphics[width=\linewidth]{img/automltime4.PNG}
%         \vspace{-0.8cm}        
%         \caption{Ensemble from SMAC algorithm on the microfossils dataset}
%         \label{fig:automltime4}
%     \end{minipage}
% \end{figure}

\subsection{step 2 - Ensemble Selection}
\label{sec:enssele}

\subsubsection{Ensemble selection pruning}
\label{sec:pruning}

We try multiple pruning factors X such X\% only the top X\% are kept. It reduces the size of the library of models and thus helps the ensemble selection algorithm. When the pruning factor is above 20\% it does not reduce accuracy and reduce the ensemble selection time, while the threshold under 15\% reduces the target metric in some experiments. For all experiments, the pruning factor is set to 20\%.

\subsubsection{Ensemble selection methods under budget}
\label{sec:ensselebench}

% Importance de l'ensemble selection
The literature \cite{useoverfit:1995} indicates the diversity is of high importance to increase ensemble machine learning. The relationship between models diversity and the accuracy of their combination is not fully understood \cite{kuncheva:2003} but the methods exploring automatically the ensemble space from a library of models and returning the most promising ensemble have been proposed \cite{caruana:2004} \cite{tsoumakas:2008} but as far we know we are the first to propose a multi-objective with a budget to suit with a heterogeneous ensemble of DNNs.

\subsubsection{SMOBF greedy compared to Forward greedy (baseline)}
\label{sec:vs}

The ensemble selection algorithm often used by most advanced AutoML software \cite{caruana:2004} \cite{autosklearn} is \textit{forward greedy with fixed number of models} that we pick as baseline. We perform three runtimes of the overall AutoML workflow and observe that SMOBF is Pareto dominant or equivalent to the baseline each time. %In the previous section this baseline could not be directly compared with our budget-constrained algorithms because the number of models does not control the computing cost.

\begin{figure}[h]
    \centering
    \begin{minipage}[]{0.45\linewidth}
        \centering
        \includegraphics[width=\linewidth]{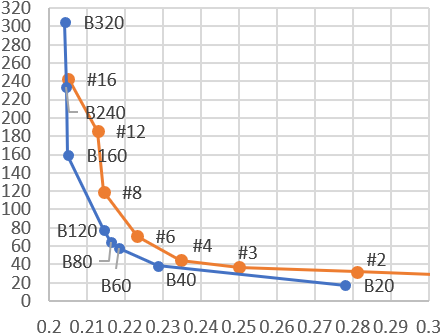}
        \vspace{-0.8cm}        
        \caption{Ensembles generated from Hyperband algorithm on the CIFAR100 dataset}
        \label{fig:greedy1}
    \end{minipage}
    \hspace*{\fill}
    % \begin{minipage}[]{0.45\linewidth}
    %     \centering
    %     \includegraphics[width=\linewidth]{img/greedy2.PNG}
    %     \vspace{-0.8cm}        
    %     \caption{Ensembles generated from SMAC algorithm on the CIFAR100 dataset}
    %     \label{fig:greedy2}
    % \end{minipage}
    % \hfill
    \begin{minipage}[]{0.45\linewidth}
        \centering
        \includegraphics[width=\linewidth]{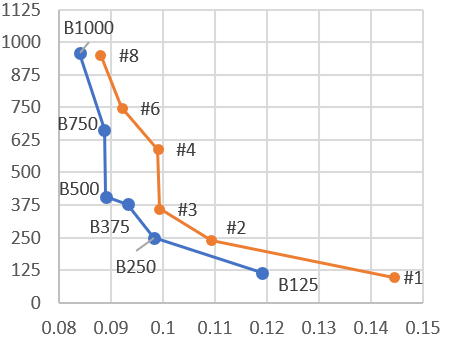}
        \vspace{-0.8cm}
        \caption{Ensembles generated from Hyperband algorithm on the microfossils dataset}
        \label{fig:greedy3}
    \end{minipage}
    \hspace*{\fill}
    % \begin{minipage}[]{0.45\linewidth}
    %     \centering
    %     \includegraphics[width=\linewidth]{img/greedy4.PNG}
    %     \vspace{-0.8cm}
    %     \caption{Ensembles generated from SMAC algorithm on the microfossils dataset}
    %     \label{fig:greedy4}
    % \end{minipage}
    %\caption{Varying the budget to construct models in function of AutoML running time. We assess two HPO algorithms (HB and SMAC) to generate models and observe the accuracy of models construction in function of the HPO running time.}
    %\label{fig:automltime}
    %\caption{Time bounded forward greedy ensemble selection with multi criterion forward greedy (orange). The mention "B20" means the budget was set to 20. We compare it with the forward greedy constrained with a fixed number of models (orange), "\#1" means that the constraint is 1 model. Horizontal axes are the accuracy error and vertical axis the cost of the ensemble, meaning the target is the bottom left point (0;0).}
\end{figure}

The figures \ref{fig:greedy1}, \ref{fig:greedy3} compares those two algorithms in function of the cost (vertically) and the error rate (horizontally) of produced ensembles with SMOBF greedy (blue) and the baseline (orange) on one runtime. On figures the mention "BX" means a budget of $X$ and "\#Y" means an ensemble of size $Y$.

The gain of SMOBF is particularly obvious when the budget is small on both criteria, but it is reduced when the budget is relaxed. Indeed, when we target the best ensemble at any cost or any size, the objective of the two algorithms converges. On the opposite when the budget is small, SMOBF informs the greedy algorithm to go toward efficient models before the budget is consumed.

\subsubsection{Assessing ensemble combiners}
\label{sec:enscomb}

During the ensemble selection process, we try three ways to best combine a candidate ensemble: majority voting, averaging, and weighted averaging (or equivalently Ensemble Selection with replacement). Results are not shown due to the lack of space but discussed.

On 6 ensembles of different sizes and different libraries of models, majority voting has shown to be inferior each time. It appears that naively averaging predictions is a simpler method and performs as well as weighted averaging them. We can explain this result by the fact that the validation information is already used during the ensemble construction, so the weights tuned on this validation set are over-fitted.

\subsection{step 3 - Ensemble allocation for serving}

Table~\ref{tab:servcifar} shows the performance in terms of prediction performance of 2 DNNs ensembles. The first one was generated with B=80 on CIFAR100 datasets it contains 7 DNNs. The second one with B=375 on the microfossils dataset containing 14 DNNs.

\begin{table}[]
\small
\centering
\begin{tabularx}{8cm}{l|ll|ll}
\toprule
\multirow{1}{*}{\#gpus} & \multicolumn{2}{l}{7 DNNs on CIFAR100} & \multicolumn{2}{l}{14 DNNs on microfossiles}  \\
(+1cpu) & algo1              & algo1+algo2              & algo1              & algo1+algo2             \\
\midrule
0          &  11 & 13           & 20                 & 28                 \\
1          & 35 & 79            & 40                 & 59                 \\
2          & 40 & 411            & 150                & 199                \\
3              & 290 & 788        & 167                & 213                \\
4               & 325 & 789        & 176                & 232                \\
5               & 375 & 791        & 198                & 244                \\
6               & 355 & 791        & 203                & 252               \\
\bottomrule
\end{tabularx}
\caption{We benchmark the predictions/seconds of an ensemble of 7 models and another of 14 models. The first one was generated with B=80 and the second one B=375}
\label{tab:servcifar}
\end{table}

% CPU MEMORY
In addition to those benchmarks, we analyzed how DNNs are allocated. We observe in both ensembles, that when we increase the number of GPUs, the CPU is quickly not used anymore. Indeed, the CPU is only used for its greater memory capacity.  We observe also that bigger models are often put alone in one GPU but smaller ones are co-localized, it is quite intuitive in terms of memory and GPU cores. Furthermore, when multiple models are in the same GPU, the batch size is chosen smaller. We observe also a bigger batch size on the CPU than on the GPU. Moreover, the second algorithm produces always a speed up but this speed is strongly dependent on the ensemble size and the available resources (from factor 10 to a few \%) confirming the usefulness of the second algorithm. The ensemble of 7 models shows that the increase of GPUs plateaus after 3 GPUs, in the second ensemble we do not see yet a plateau with 6 GPUs. Those results teach us that to serve efficiently an ensemble of DNNs we do not need systematically as much as GPUs as DNNs.

%Due to lack of time, we do not analyze if the speed up of the algorithm~\ref{algo:speed} comes from the refinement of batch size values or the refinement of DNNs placement.

\section{Baseline comparison}

We report in figure~\ref{tab:comparautoml} results of authors running until convergence multiple AutoML methods with diverse theoretical backgrounds \cite{nas:2020}, \cite{evolution:2017}, \cite{evolution:2020}. We run the Auto-Keras framework version 1.0.12 with one single GPU and default settings, except the max\_trials set to 10. We notice no improvement when max\_trials is set to 20. We report our proposed workflow results and some intermediate scores over time.

% 2 trends
The comparison between AutoML methods is known to be difficult because authors explore different search spaces, with different initial conditions, different data processing, and ensembles, however, we observe two main trends. The first block of rows AutoML methods converging fast in several GPU hours which often explore the neural architecture space sequentially on one GPU. The second block is evolution-based algorithms that converge later, their population-based approach allows to easily leverage GPU clusters. Asynchronous Hyperband is still not widely used.

% Hyperband less sensitive to settings
As evolutionary methods, our method can benefit from high computing power. In comparison, asynchronous Hyperband does not require intermediate stopping of all GPUs which makes it more suitable to use GPU clusters compared to a sequence of generations. Also, in the large-scale case where the number of available GPUs is superior to the number of trials, Hyperband can run all trials at the same time, while genetics is limited by running all trials of the current generation.

Furthermore, Hyperband has low settings requirements.  It uses early stopping which reduces the sensibility to the \#trials/\#epochs dilemma. Also, the exploration/exploitation of the hyperparameter space is balanced by stopping a fraction of less promising trials, so only the halving factor is needed. Evolutionary methods require much more initial settings such as the number of generations, the crossover operation, the mutation operation. They make this algorithm sensitive to the initial choices and so the settings calibrated for an application cannot be suited for a new application.

We try our best to do a fair comparison by using the same dataset. However, different data processing, different hardware, different initial settings can have an important impact on experimental results. We do not perform cutout and yet it seems effective processing on CIFAR100. Finally, we recall that with the given time horizon of 36GPU/hours visible in figure~\ref{fig:greedy1} and figure~\ref{fig:greedy3}, Hyperband+SMOB greedy has not yet converged.

\begin{table}[]
%\addtolength{\tabcolsep}{-4pt}
\small
\centering
\begin{tabularx}{11cm}{l|cccc|c|c}
\toprule
Method                     & \#GPU & hours & Cum h & GPU name & \#DNNs & Test(\%)             \\
\midrule
RSPS \cite{rsps:2020}                                                 & 1 & 2 & 2                                                      &  GTX1080TI  & 1 & 52.31$\pm$5.77   \\
DARTS-V1 \cite{darts:2018}                                               & 1 & 3 & 3                                                       &  GTX1080TI  & 1 & 15.03$\pm$0.00   \\
DARTS-V2                                               & 1 & 10  & 10                                                        &  GTX1080TI & 1 & 15.03$\pm$0.00 \\
GDAS  \cite{gdas:2019}                                                 & 1 & 9 & 9                                                       &  GTX1080TI  & 1 & 71.34$\pm$0.34  \\
SETN  \cite{setn:2019}                                                 & 1 & 10 & 10                                                      &  GTX1080TI & 1  & 58.86$\pm$0.06  \\
ENAS   \cite{ENAS:2018}                                                & 1 & 4 & 4                                                       &  GTX1080TI & 1  & 13.37$\pm$2.35  \\
Auto-Keras \cite{autokeras}                                        & 1 & 2 & 2                                                        & Tesla V100 & 1 & 69.57$\pm$0.53 \\
\hline 
LSE \cite{evolution:2017}  & 250 & 264 & 66K             & ? & ? & 77.0              \\
CNN-GA \cite{evolution:2020} & 3 & 320 & 960                                         &  GTX1080TI         & 1     & 77.97             \\
CNN-GA+cutout \cite{evolution:2020} & 3 & 320 & 960                                         &  GTX1080TI         & 1     & 79.47             \\
\hline 
Ours with B=320                                      & 6 & 6 & 36         & Tesla V100       & 4 & 72.39  \\
Ours with B=320                                       & 6 & 24 & 144      & Tesla V100       & 18 & 77.35  \\
Ours with B=320                                       & 6 & 144  & 864    & Tesla V100       & 34 & 79.44   \\
\bottomrule
\end{tabularx}
\caption{The comparison between AutoML algorithms in terms of the classification accuracy (\%) and GPU hours on CIFAR100 benchmark dataset. The "?" mention means the information is missing in the paper. Column from left to right are: the name of the method, the number of GPUs used, duration of the algorithm (hours), cumulated time, GPU name, number of models (1=no ensembling) , mean test accuracy}
\label{tab:comparautoml}
\end{table}

\section{Future works}

%The bottleneck performance of our deployed server is still the neural network prediction computing cost, it limits the number of assessed combinations and the speed of those combinations.

% Therefore its deployment file format and numeric representation have an impact on the overall performance. Today, we use the ".pb" file format which is both compatible with GPUs and CPUs. Some other formats could again increase performance but are hardware-specific such as OpenVINO \cite{openvino} and TensorRT \cite{tensorrt}.

% Post-training quantization is also a simple trick to speed up the inference and reduce its memory footprint with a low impact on accuracy.

%With the given server architecture changing the file format has an isolated impact on the code of the \textit{DNN predictor}.

%There are generally now two ways to
%speed up an ensemble of DNN: training a simpler model with fewer parameters,
%or simplifying the original model using approximation techniques
%such as matrix factorization, pruning, and quantization

%We propose a pipeline to build an efficient ensemble of DNNs running a  multi-objective Ensemble Selection procedure with budget constraint (SMOB) followed by a novel efficient inference system for DNNs. 
To speed up the inference service of an ensemble, they are two ways, either running SMOBF with a lower budget to build another light ensemble or running post-training network optimization on each DNN. Post-training optimization is an active field of research, such as weights pruning \cite{pruning:2017} and weights quantization \cite{compression:2016}, \cite{compressentropy:2020}. All these methods may have a  low or no impact on the accuracy.

In those experiments, we use the Tensorflow inference engine (".pb" file format), which is both compatible with GPUs and CPUs. Some inference engine frameworks perform post-training optimization and platform-specific optimizations such as OpenVINO \cite{openvino}  for Intel CPUs and TensorRT \cite{tensorrt} for Nvidia GPUs.

Those lines of research should again increase the effectiveness and popularize the automatic construction of a heterogeneous ensemble of DNNs with a smart allocation strategy.

\section*{Conclusion}

Due to the increasing number of new Deep Learning applications and datasets, Auto Machine Learning (AutoML) methods are an important line of research. We propose an AutoML workflow capable to tune, train, ensembling, and deploy DNNs automatically but that runs a heavy workload at each stage. We aim to fill the gap between Machine Learning researches, the new GPU clusters, and the end-user application quality of service. To go toward this direction, we formulate the problems by aiming at the accuracy, the inference speed, and the flexibility of the underlying heterogeneous infrastructure.

First, we presented the experimental results demonstrating that asynchronous Hyperband is suitable for parallelism and generates the best library of models to ensemble them. We then propose a new Ensemble Selection strategy that allows controlling the final ensemble computing cost of heterogenous DNNs. When the budget is relaxed, our algorithm offers high and robust accuracy compared to other AutoML workflows. Finally, we propose a solution to the complex allocation problem of DNNs into GPUs to democratize heterogeneous ensembles even if the number of DNNs is larger than the number of GPUs. % It is conceived with a flexibility never obtained before. %, the ensemble leverage of high computing power with data-parallelism but it is also suitable for lower-cost servers, with co-localization. It is also applicable to multiple application, the code update are well identified in the \textit{pre-processor}, \textit{post-processor} and the \textit{prediction accumulator} thread classes. Future DNN representation can be pushed in the \textit{DNN predictor} thread class to again improve performance.

The history of Machine Learning is correlated to the available computing power. Since the emergence of multi-core processors in the 2000s allowed to stride from simple statistical models (e.g., decision tree) to machine learning based on ensemble (e.g., Random Forest). Then, GPGPU allows to stride from non-deep machine learning using a few cores to deep learning using hundreds of cores. GPU clusters are undoubtedly the dawn of a new era for future deep learning methods such as AutoML with ensembling.

\subsubsection*{Acknowledgement}

We would like to thank TotalEnergies SE and its subsidiaries for allowing us to share this material and make available the needed resources.

\bibliographystyle{ACM-Reference-Format}
\bibliography{bib_automl, bib_biodiv, bib_dl, bib_db, bib_ensemble, bib_hpml, bib_infer, bib_rl}


\begin{thebibliography}{59}

%%% the Web address.

\ifx \showCODEN    \undefined \def \showCODEN     #1{\unskip}     \fi
\ifx \showDOI      \undefined \def \showDOI       #1{#1}\fi
\ifx \showISBNx    \undefined \def \showISBNx     #1{\unskip}     \fi
\ifx \showISBNxiii \undefined \def \showISBNxiii  #1{\unskip}     \fi
\ifx \showISSN     \undefined \def \showISSN      #1{\unskip}     \fi
\ifx \showLCCN     \undefined \def \showLCCN      #1{\unskip}     \fi
\ifx \shownote     \undefined \def \shownote      #1{#1}          \fi
\ifx \showarticletitle \undefined \def \showarticletitle #1{#1}   \fi
\ifx \showURL      \undefined \def \showURL       {\relax}        \fi
% invisible to TeX
\providecommand\bibfield[2]{#2}
\providecommand\bibinfo[2]{#2}
\providecommand\natexlab[1]{#1}
\providecommand\showeprint[2][]{arXiv:#2}

\bibitem[\protect\citeauthoryear{Abadi, Barham, Chen, Chen, Davis, Dean, Devin,
  Ghemawat, Irving, Isard, Kudlur, Levenberg, Monga, Moore, Murray, Steiner,
  Tucker, Vasudevan, Warden, Wicke, Yu, and Zheng}{Abadi et~al\mbox{.}}{2016}]%
        {tf:2020}
\bibfield{author}{\bibinfo{person}{Mart\'{\i}n Abadi}, \bibinfo{person}{Paul
  Barham}, \bibinfo{person}{Jianmin Chen}, \bibinfo{person}{Zhifeng Chen},
  \bibinfo{person}{Andy Davis}, \bibinfo{person}{Jeffrey Dean},
  \bibinfo{person}{Matthieu Devin}, \bibinfo{person}{Sanjay Ghemawat},
  \bibinfo{person}{Geoffrey Irving}, \bibinfo{person}{Michael Isard},
  \bibinfo{person}{Manjunath Kudlur}, \bibinfo{person}{Josh Levenberg},
  \bibinfo{person}{Rajat Monga}, \bibinfo{person}{Sherry Moore},
  \bibinfo{person}{Derek~G. Murray}, \bibinfo{person}{Benoit Steiner},
  \bibinfo{person}{Paul Tucker}, \bibinfo{person}{Vijay Vasudevan},
  \bibinfo{person}{Pete Warden}, \bibinfo{person}{Martin Wicke},
  \bibinfo{person}{Yuan Yu}, {and} \bibinfo{person}{Xiaoqiang Zheng}.}
  \bibinfo{year}{2016}\natexlab{}.
\newblock \showarticletitle{TensorFlow: A System for Large-Scale Machine
  Learning}. In \bibinfo{booktitle}{\emph{Proceedings of the 12th USENIX
  Conference on Operating Systems Design and Implementation}} (Savannah, GA,
  USA) \emph{(\bibinfo{series}{OSDI'16})}. \bibinfo{publisher}{USENIX
  Association}, \bibinfo{address}{USA}, \bibinfo{pages}{265–283}.
\newblock
\showISBNx{9781931971331}


\bibitem[\protect\citeauthoryear{Bergstra and Bengio}{Bergstra and
  Bengio}{2012}]%
        {rs:2012}
\bibfield{author}{\bibinfo{person}{James Bergstra} {and}
  \bibinfo{person}{Yoshua Bengio}.} \bibinfo{year}{2012}\natexlab{}.
\newblock \showarticletitle{Random Search for Hyper-Parameter Optimization.}
\newblock \bibinfo{journal}{\emph{J. Mach. Learn. Res.}}  \bibinfo{volume}{13}
  (\bibinfo{year}{2012}), \bibinfo{pages}{281--305}.
\newblock
\urldef\tempurl%
\url{http://dblp.uni-trier.de/db/journals/jmlr/jmlr13.html#BergstraB12}
\showURL{%
\tempurl}


\bibitem[\protect\citeauthoryear{Bergstra, Bardenet, Bengio, and
  K\'{e}gl}{Bergstra et~al\mbox{.}}{2011}]%
        {TPE:2011}
\bibfield{author}{\bibinfo{person}{James~S. Bergstra},
  \bibinfo{person}{R\'{e}mi Bardenet}, \bibinfo{person}{Yoshua Bengio}, {and}
  \bibinfo{person}{Bal\'{a}zs K\'{e}gl}.} \bibinfo{year}{2011}\natexlab{}.
\newblock \showarticletitle{Algorithms for Hyper-Parameter Optimization}.
\newblock In \bibinfo{booktitle}{\emph{Advances in Neural Information
  Processing Systems 24}}, \bibfield{editor}{\bibinfo{person}{J.~Shawe-Taylor},
  \bibinfo{person}{R.~S. Zemel}, \bibinfo{person}{P.~L. Bartlett},
  \bibinfo{person}{F.~Pereira}, {and} \bibinfo{person}{K.~Q. Weinberger}}
  (Eds.). \bibinfo{publisher}{Curran Associates, Inc.},
  \bibinfo{pages}{2546--2554}.
\newblock
\urldef\tempurl%
\url{http://papers.nips.cc/paper/4443-algorithms-for-hyper-parameter-optimization.pdf}
\showURL{%
\tempurl}


\bibitem[\protect\citeauthoryear{Brown, Wyatt, Harris, and Yao}{Brown
  et~al\mbox{.}}{2005}]%
        {creatediversity:2005}
\bibfield{author}{\bibinfo{person}{Gavin Brown}, \bibinfo{person}{Jeremy
  Wyatt}, \bibinfo{person}{Rachel Harris}, {and} \bibinfo{person}{Xin Yao}.}
  \bibinfo{year}{2005}\natexlab{}.
\newblock \showarticletitle{Diversity Creation Methods: A Survey And
  Categorisation}.
\newblock \bibinfo{journal}{\emph{Information Fusion}}  \bibinfo{volume}{6}
  (\bibinfo{date}{03} \bibinfo{year}{2005}), \bibinfo{pages}{5--20}.
\newblock
\urldef\tempurl%
\url{https://doi.org/10.1016/j.inffus.2004.04.004}
\showDOI{\tempurl}


\bibitem[\protect\citeauthoryear{Caruana, Munson, and Niculescu-Mizil}{Caruana
  et~al\mbox{.}}{2006}]%
        {Caruana:2006}
\bibfield{author}{\bibinfo{person}{Rich Caruana}, \bibinfo{person}{Art Munson},
  {and} \bibinfo{person}{Alexandru Niculescu-Mizil}.}
  \bibinfo{year}{2006}\natexlab{}.
\newblock \showarticletitle{Getting the Most Out of Ensemble Selection}.
\newblock \bibinfo{journal}{\emph{Proceedings - IEEE International Conference
  on Data Mining, ICDM}}, \bibinfo{pages}{828--833}.
\newblock
\urldef\tempurl%
\url{https://doi.org/10.1109/ICDM.2006.76}
\showDOI{\tempurl}


\bibitem[\protect\citeauthoryear{Caruana, Niculescu-Mizil, Crew, and
  Ksikes}{Caruana et~al\mbox{.}}{2004}]%
        {caruana:2004}
\bibfield{author}{\bibinfo{person}{Rich Caruana}, \bibinfo{person}{Alexandru
  Niculescu-Mizil}, \bibinfo{person}{Geoff Crew}, {and} \bibinfo{person}{Alex
  Ksikes}.} \bibinfo{year}{2004}\natexlab{}.
\newblock \showarticletitle{Ensemble Selection from Libraries of Models}. In
  \bibinfo{booktitle}{\emph{Proceedings of the Twenty-First International
  Conference on Machine Learning}} (Banff, Alberta, Canada)
  \emph{(\bibinfo{series}{ICML '04})}. \bibinfo{publisher}{Association for
  Computing Machinery}, \bibinfo{address}{New York, NY, USA},
  \bibinfo{pages}{18}.
\newblock
\showISBNx{1581138385}
\urldef\tempurl%
\url{https://doi.org/10.1145/1015330.1015432}
\showDOI{\tempurl}


\bibitem[\protect\citeauthoryear{Cea, Gruen, and Richmond}{Cea
  et~al\mbox{.}}{2021}]%
        {ensimg:2021}
\bibfield{author}{\bibinfo{person}{María V. Sainz~de Cea},
  \bibinfo{person}{David Gruen}, {and} \bibinfo{person}{David Richmond}.}
  \bibinfo{year}{2021}\natexlab{}.
\newblock \showarticletitle{Pneumoperitoneum Detection In Chest X-Ray By A Deep
  Learning Ensemble With Model Explainability}. In
  \bibinfo{booktitle}{\emph{2021 IEEE 18th International Symposium on
  Biomedical Imaging (ISBI)}}. \bibinfo{pages}{1637--1641}.
\newblock
\urldef\tempurl%
\url{https://doi.org/10.1109/ISBI48211.2021.9434122}
\showDOI{\tempurl}


\bibitem[\protect\citeauthoryear{Chen, Wu, Mo, Chattopadhyay, and Lipson}{Chen
  et~al\mbox{.}}{2018}]%
        {autostacker}
\bibfield{author}{\bibinfo{person}{Boyuan Chen}, \bibinfo{person}{Harvey Wu},
  \bibinfo{person}{Warren Mo}, \bibinfo{person}{Ishanu Chattopadhyay}, {and}
  \bibinfo{person}{Hod Lipson}.} \bibinfo{year}{2018}\natexlab{}.
\newblock \showarticletitle{Autostacker: A Compositional Evolutionary Learning
  System}. In \bibinfo{booktitle}{\emph{Proceedings of the Genetic and
  Evolutionary Computation Conference}} (Kyoto, Japan)
  \emph{(\bibinfo{series}{GECCO '18})}. \bibinfo{publisher}{Association for
  Computing Machinery}, \bibinfo{address}{New York, NY, USA},
  \bibinfo{pages}{402–409}.
\newblock
\showISBNx{9781450356183}
\urldef\tempurl%
\url{https://doi.org/10.1145/3205455.3205586}
\showDOI{\tempurl}


\bibitem[\protect\citeauthoryear{Chollet}{Chollet}{2017}]%
        {Xception:2016}
\bibfield{author}{\bibinfo{person}{François Chollet}.}
  \bibinfo{year}{2017}\natexlab{}.
\newblock \showarticletitle{Xception: Deep Learning with Depthwise Separable
  Convolutions}.
\newblock \bibinfo{journal}{\emph{2017 IEEE Conference on Computer Vision and
  Pattern Recognition (CVPR)}} (\bibinfo{year}{2017}),
  \bibinfo{pages}{1800--1807}.
\newblock


\bibitem[\protect\citeauthoryear{Cordts, Omran, Ramos, Rehfeld, Enzweiler,
  Benenson, Franke, Roth, and Schiele}{Cordts et~al\mbox{.}}{2016}]%
        {citydb:2016}
\bibfield{author}{\bibinfo{person}{Marius Cordts}, \bibinfo{person}{Mohamed
  Omran}, \bibinfo{person}{Sebastian Ramos}, \bibinfo{person}{Timo Rehfeld},
  \bibinfo{person}{Markus Enzweiler}, \bibinfo{person}{Rodrigo Benenson},
  \bibinfo{person}{Uwe Franke}, \bibinfo{person}{Stefan Roth}, {and}
  \bibinfo{person}{Bernt Schiele}.} \bibinfo{year}{2016}\natexlab{}.
\newblock \showarticletitle{The Cityscapes Dataset for Semantic Urban Scene
  Understanding}. In \bibinfo{booktitle}{\emph{Proc. of the IEEE Conference on
  Computer Vision and Pattern Recognition (CVPR)}}.
\newblock


\bibitem[\protect\citeauthoryear{Davoodi, Gwon, Lai, and Morris}{Davoodi
  et~al\mbox{.}}{2019}]%
        {tensorrt}
\bibfield{author}{\bibinfo{person}{Pooya Davoodi}, \bibinfo{person}{Chul Gwon},
  \bibinfo{person}{Guangda Lai}, {and} \bibinfo{person}{Trevor Morris}.}
  \bibinfo{year}{2019}\natexlab{}.
\newblock \showarticletitle{“TensorRT inference With TensorFlow”}.
\newblock
\newblock
\shownote{GPU Technology Conference.}


\bibitem[\protect\citeauthoryear{Dong and Yang}{Dong and Yang}{2019a}]%
        {setn:2019}
\bibfield{author}{\bibinfo{person}{Xuanyi Dong} {and} \bibinfo{person}{Yi
  Yang}.} \bibinfo{year}{2019}\natexlab{a}.
\newblock \showarticletitle{One-Shot Neural Architecture Search via
  Self-Evaluated Template Network}. In \bibinfo{booktitle}{\emph{Proceedings of
  the IEEE/CVF International Conference on Computer Vision (ICCV)}}.
\newblock


\bibitem[\protect\citeauthoryear{Dong and Yang}{Dong and Yang}{2019b}]%
        {gdas:2019}
\bibfield{author}{\bibinfo{person}{Xuanyi Dong} {and} \bibinfo{person}{Yi
  Yang}.} \bibinfo{year}{2019}\natexlab{b}.
\newblock \showarticletitle{Searching for a Robust Neural Architecture in Four
  GPU Hours}. In \bibinfo{booktitle}{\emph{Proceedings of the IEEE/CVF
  Conference on Computer Vision and Pattern Recognition (CVPR)}}.
\newblock


\bibitem[\protect\citeauthoryear{Feurer, Klein, Eggensperger, Springenberg,
  Blum, and Hutter}{Feurer et~al\mbox{.}}{2015}]%
        {autosklearn}
\bibfield{author}{\bibinfo{person}{Matthias Feurer}, \bibinfo{person}{Aaron
  Klein}, \bibinfo{person}{Katharina Eggensperger}, \bibinfo{person}{Jost
  Springenberg}, \bibinfo{person}{Manuel Blum}, {and} \bibinfo{person}{Frank
  Hutter}.} \bibinfo{year}{2015}\natexlab{}.
\newblock \showarticletitle{Efficient and Robust Automated Machine Learning}.
\newblock In \bibinfo{booktitle}{\emph{Advances in Neural Information
  Processing Systems 28}}, \bibfield{editor}{\bibinfo{person}{C.~Cortes},
  \bibinfo{person}{N.~D. Lawrence}, \bibinfo{person}{D.~D. Lee},
  \bibinfo{person}{M.~Sugiyama}, {and} \bibinfo{person}{R.~Garnett}} (Eds.).
  \bibinfo{publisher}{Curran Associates, Inc.}, \bibinfo{pages}{2962--2970}.
\newblock
\urldef\tempurl%
\url{http://papers.nips.cc/paper/5872-efficient-and-robust-automated-machine-learning.pdf}
\showURL{%
\tempurl}


\bibitem[\protect\citeauthoryear{Ge and Jones}{Ge and Jones}{2018}]%
        {openvino}
\bibfield{author}{\bibinfo{person}{Yi Ge} {and} \bibinfo{person}{Monique
  Jones}.} \bibinfo{year}{2018}\natexlab{}.
\newblock \showarticletitle{Inference With Intel}.
\newblock
\newblock
\shownote{AI DevCon 2018.}


\bibitem[\protect\citeauthoryear{Guan, Mokadam, Shen, Lim, and Patton}{Guan
  et~al\mbox{.}}{2020}]%
        {fleet}
\bibfield{author}{\bibinfo{person}{Hui Guan}, \bibinfo{person}{Laxmikant~Kishor
  Mokadam}, \bibinfo{person}{Xipeng Shen}, \bibinfo{person}{Seung-Hwan Lim},
  {and} \bibinfo{person}{Robert Patton}.} \bibinfo{year}{2020}\natexlab{}.
\newblock \showarticletitle{FLEET: Flexible Efficient Ensemble Training for
  Heterogeneous Deep Neural Networks}. In \bibinfo{booktitle}{\emph{Proceedings
  of Machine Learning and Systems}},
  \bibfield{editor}{\bibinfo{person}{I.~Dhillon},
  \bibinfo{person}{D.~Papailiopoulos}, {and} \bibinfo{person}{V.~Sze}} (Eds.),
  Vol.~\bibinfo{volume}{2}. \bibinfo{pages}{247--261}.
\newblock
\urldef\tempurl%
\url{https://proceedings.mlsys.org/paper/2020/file/ed3d2c21991e3bef5e069713af9fa6ca-Paper.pdf}
\showURL{%
\tempurl}


\bibitem[\protect\citeauthoryear{Gulli and Pal}{Gulli and Pal}{2017}]%
        {keras:2017}
\bibfield{author}{\bibinfo{person}{Antonio Gulli} {and} \bibinfo{person}{Sujit
  Pal}.} \bibinfo{year}{2017}\natexlab{}.
\newblock \bibinfo{booktitle}{\emph{Deep learning with Keras}}.
\newblock \bibinfo{publisher}{Packt Publishing Ltd}.
\newblock


\bibitem[\protect\citeauthoryear{Guyon, Sun-Hosoya, Boull{\'e}, Escalante,
  Escalera, Liu, Jajetic, Ray, Saeed, Sebag, Statnikov, Tu, and Viegas}{Guyon
  et~al\mbox{.}}{2019}]%
        {Guyon2019}
\bibfield{author}{\bibinfo{person}{Isabelle Guyon}, \bibinfo{person}{Lisheng
  Sun-Hosoya}, \bibinfo{person}{Marc Boull{\'e}}, \bibinfo{person}{Hugo~Jair
  Escalante}, \bibinfo{person}{Sergio Escalera}, \bibinfo{person}{Zhengying
  Liu}, \bibinfo{person}{Damir Jajetic}, \bibinfo{person}{Bisakha Ray},
  \bibinfo{person}{Mehreen Saeed}, \bibinfo{person}{Mich{\`e}le Sebag},
  \bibinfo{person}{Alexander Statnikov}, \bibinfo{person}{Wei-Wei Tu}, {and}
  \bibinfo{person}{Evelyne Viegas}.} \bibinfo{year}{2019}\natexlab{}.
\newblock \bibinfo{booktitle}{\emph{Analysis of the AutoML Challenge Series
  2015--2018}}.
\newblock \bibinfo{publisher}{Springer International Publishing},
  \bibinfo{address}{Cham}, \bibinfo{pages}{177--219}.
\newblock
\showISBNx{978-3-030-05318-5}
\urldef\tempurl%
\url{https://doi.org/10.1007/978-3-030-05318-5_10}
\showDOI{\tempurl}


\bibitem[\protect\citeauthoryear{Haider, Akhunzada, Mustafa, Patel, Fernandez,
  Choo, and Iqbal}{Haider et~al\mbox{.}}{2020}]%
        {enscyb:2020}
\bibfield{author}{\bibinfo{person}{Shahzeb Haider}, \bibinfo{person}{Adnan
  Akhunzada}, \bibinfo{person}{Iqra Mustafa}, \bibinfo{person}{Tanil~Bharat
  Patel}, \bibinfo{person}{Amanda Fernandez},
  \bibinfo{person}{Kim-Kwang~Raymond Choo}, {and} \bibinfo{person}{Javed
  Iqbal}.} \bibinfo{year}{2020}\natexlab{}.
\newblock \showarticletitle{A Deep CNN Ensemble Framework for Efficient DDoS
  Attack Detection in Software Defined Networks}.
\newblock \bibinfo{journal}{\emph{IEEE Access}}  \bibinfo{volume}{8}
  (\bibinfo{year}{2020}), \bibinfo{pages}{53972--53983}.
\newblock
\urldef\tempurl%
\url{https://doi.org/10.1109/ACCESS.2020.2976908}
\showDOI{\tempurl}


\bibitem[\protect\citeauthoryear{Han, Mao, and Dally}{Han
  et~al\mbox{.}}{2016}]%
        {compression:2016}
\bibfield{author}{\bibinfo{person}{Song Han}, \bibinfo{person}{Huizi Mao},
  {and} \bibinfo{person}{William~J. Dally}.} \bibinfo{year}{2016}\natexlab{}.
\newblock \showarticletitle{Deep Compression: Compressing Deep Neural Network
  with Pruning, Trained Quantization and Huffman Coding}. In
  \bibinfo{booktitle}{\emph{4th International Conference on Learning
  Representations, {ICLR} 2016, San Juan, Puerto Rico, May 2-4, 2016,
  Conference Track Proceedings}}, \bibfield{editor}{\bibinfo{person}{Yoshua
  Bengio} {and} \bibinfo{person}{Yann LeCun}} (Eds.).
\newblock


\bibitem[\protect\citeauthoryear{He, Zhang, Ren, and Sun}{He
  et~al\mbox{.}}{2016}]%
        {resnet:2015}
\bibfield{author}{\bibinfo{person}{Kaiming He}, \bibinfo{person}{Xiangyu
  Zhang}, \bibinfo{person}{Shaoqing Ren}, {and} \bibinfo{person}{Jian Sun}.}
  \bibinfo{year}{2016}\natexlab{}.
\newblock \showarticletitle{Deep Residual Learning for Image Recognition}. In
  \bibinfo{booktitle}{\emph{2016 IEEE Conference on Computer Vision and Pattern
  Recognition (CVPR)}}. \bibinfo{pages}{770--778}.
\newblock
\urldef\tempurl%
\url{https://doi.org/10.1109/CVPR.2016.90}
\showDOI{\tempurl}


\bibitem[\protect\citeauthoryear{Hoffman, Brochu, and de~Freitas}{Hoffman
  et~al\mbox{.}}{2011}]%
        {bogp:2011}
\bibfield{author}{\bibinfo{person}{Matthew Hoffman}, \bibinfo{person}{Eric
  Brochu}, {and} \bibinfo{person}{Nando de Freitas}.}
  \bibinfo{year}{2011}\natexlab{}.
\newblock \showarticletitle{Portfolio Allocation for Bayesian Optimization}. In
  \bibinfo{booktitle}{\emph{Proceedings of the Twenty-Seventh Conference on
  Uncertainty in Artificial Intelligence}} (Barcelona, Spain)
  \emph{(\bibinfo{series}{UAI'11})}. \bibinfo{publisher}{AUAI Press},
  \bibinfo{address}{Arlington, Virginia, USA}, \bibinfo{pages}{327–336}.
\newblock
\showISBNx{9780974903972}


\bibitem[\protect\citeauthoryear{Huang, Liu, Van Der~Maaten, and
  Weinberger}{Huang et~al\mbox{.}}{2017}]%
        {densenet:2016}
\bibfield{author}{\bibinfo{person}{Gao Huang}, \bibinfo{person}{Zhuang Liu},
  \bibinfo{person}{Laurens Van Der~Maaten}, {and} \bibinfo{person}{Kilian~Q.
  Weinberger}.} \bibinfo{year}{2017}\natexlab{}.
\newblock \showarticletitle{Densely Connected Convolutional Networks}. In
  \bibinfo{booktitle}{\emph{2017 IEEE Conference on Computer Vision and Pattern
  Recognition (CVPR)}}. \bibinfo{pages}{2261--2269}.
\newblock
\urldef\tempurl%
\url{https://doi.org/10.1109/CVPR.2017.243}
\showDOI{\tempurl}


\bibitem[\protect\citeauthoryear{Hutter, Hoos, and Leyton-Brown}{Hutter
  et~al\mbox{.}}{2011}]%
        {SMAC}
\bibfield{author}{\bibinfo{person}{Frank Hutter}, \bibinfo{person}{Holger~H
  Hoos}, {and} \bibinfo{person}{Kevin Leyton-Brown}.}
  \bibinfo{year}{2011}\natexlab{}.
\newblock \showarticletitle{Sequential model-based optimization for general
  algorithm configuration}. In \bibinfo{booktitle}{\emph{International
  conference on learning and intelligent optimization}}. Springer,
  \bibinfo{pages}{507--523}.
\newblock


\bibitem[\protect\citeauthoryear{Jin, Song, and Hu}{Jin et~al\mbox{.}}{2019}]%
        {autokeras}
\bibfield{author}{\bibinfo{person}{Haifeng Jin}, \bibinfo{person}{Qingquan
  Song}, {and} \bibinfo{person}{Xia Hu}.} \bibinfo{year}{2019}\natexlab{}.
\newblock \showarticletitle{Auto-Keras: An Efficient Neural Architecture Search
  System}. In \bibinfo{booktitle}{\emph{Proceedings of the 25th ACM SIGKDD
  International Conference on Knowledge Discovery and Data Mining}} (Anchorage,
  AK, USA) \emph{(\bibinfo{series}{KDD '19})}. \bibinfo{publisher}{Association
  for Computing Machinery}, \bibinfo{address}{New York, NY, USA},
  \bibinfo{pages}{1946–1956}.
\newblock
\showISBNx{9781450362016}
\urldef\tempurl%
\url{https://doi.org/10.1145/3292500.3330648}
\showDOI{\tempurl}


\bibitem[\protect\citeauthoryear{Johnston, Young, Hughes, Patton, and
  White}{Johnston et~al\mbox{.}}{2017}]%
        {cloud:2017}
\bibfield{author}{\bibinfo{person}{Travis Johnston}, \bibinfo{person}{Steven~R.
  Young}, \bibinfo{person}{David Hughes}, \bibinfo{person}{Robert~M. Patton},
  {and} \bibinfo{person}{Devin White}.} \bibinfo{year}{2017}\natexlab{}.
\newblock \showarticletitle{Optimizing Convolutional Neural Networks for Cloud
  Detection}. In \bibinfo{booktitle}{\emph{Proceedings of the Machine Learning
  on HPC Environments}} (Denver, CO, USA) \emph{(\bibinfo{series}{MLHPC'17})}.
  \bibinfo{publisher}{Association for Computing Machinery},
  \bibinfo{address}{New York, NY, USA}, Article \bibinfo{articleno}{4},
  \bibinfo{numpages}{9}~pages.
\newblock
\showISBNx{9781450351379}
\urldef\tempurl%
\url{https://doi.org/10.1145/3146347.3146352}
\showDOI{\tempurl}


\bibitem[\protect\citeauthoryear{Kingma and Ba}{Kingma and Ba}{2014}]%
        {adam:2014}
\bibfield{author}{\bibinfo{person}{Diederik Kingma} {and}
  \bibinfo{person}{Jimmy Ba}.} \bibinfo{year}{2014}\natexlab{}.
\newblock \showarticletitle{Adam: A Method for Stochastic Optimization}.
\newblock \bibinfo{journal}{\emph{International Conference on Learning
  Representations}} (\bibinfo{date}{12} \bibinfo{year}{2014}).
\newblock


\bibitem[\protect\citeauthoryear{Krizhevsky}{Krizhevsky}{2009}]%
        {cifardb:2009}
\bibfield{author}{\bibinfo{person}{Alex Krizhevsky}.}
  \bibinfo{year}{2009}\natexlab{}.
\newblock \bibinfo{booktitle}{\emph{Learning multiple layers of features from
  tiny images}}.
\newblock \bibinfo{type}{{T}echnical {R}eport}.
\newblock


\bibitem[\protect\citeauthoryear{Kuncheva and Whitaker}{Kuncheva and
  Whitaker}{2003}]%
        {kuncheva:2003}
\bibfield{author}{\bibinfo{person}{Ludmila Kuncheva} {and}
  \bibinfo{person}{Chris Whitaker}.} \bibinfo{year}{2003}\natexlab{}.
\newblock \showarticletitle{Measures of Diversity in Classifier Ensembles and
  Their Relationship with the Ensemble Accuracy}.
\newblock \bibinfo{journal}{\emph{Machine Learning}}  \bibinfo{volume}{51}
  (\bibinfo{date}{05} \bibinfo{year}{2003}), \bibinfo{pages}{181--207}.
\newblock
\urldef\tempurl%
\url{https://doi.org/10.1023/A:1022859003006}
\showDOI{\tempurl}


\bibitem[\protect\citeauthoryear{LeDell and Poirier}{LeDell and
  Poirier}{2020}]%
        {h2oautoml}
\bibfield{author}{\bibinfo{person}{Erin LeDell} {and}
  \bibinfo{person}{Sebastien Poirier}.} \bibinfo{year}{2020}\natexlab{}.
\newblock \showarticletitle{{H2O} {A}uto{ML}: Scalable Automatic Machine
  Learning}.
\newblock \bibinfo{journal}{\emph{7th ICML Workshop on Automated Machine
  Learning (AutoML)}} (\bibinfo{date}{July} \bibinfo{year}{2020}).
\newblock
\urldef\tempurl%
\url{https://www.automl.org/wp-content/uploads/2020/07/AutoML_2020_paper_61.pdf}
\showURL{%
\tempurl}


\bibitem[\protect\citeauthoryear{Li, Yu, Fu, Zhang, Zhao, You, Yu, Wang, Hao,
  and Lin}{Li et~al\mbox{.}}{2021b}]%
        {nas:2020}
\bibfield{author}{\bibinfo{person}{Chaojian Li}, \bibinfo{person}{Zhongzhi Yu},
  \bibinfo{person}{Yonggan Fu}, \bibinfo{person}{Yongan Zhang},
  \bibinfo{person}{Yang Zhao}, \bibinfo{person}{Haoran You},
  \bibinfo{person}{Qixuan Yu}, \bibinfo{person}{Yue Wang},
  \bibinfo{person}{Cong Hao}, {and} \bibinfo{person}{Yingyan Lin}.}
  \bibinfo{year}{2021}\natexlab{b}.
\newblock \showarticletitle{{\{}HW{\}}-{\{}NAS{\}}-Bench: Hardware-Aware Neural
  Architecture Search Benchmark}. In \bibinfo{booktitle}{\emph{International
  Conference on Learning Representations}}.
\newblock
\urldef\tempurl%
\url{https://openreview.net/forum?id=_0kaDkv3dVf}
\showURL{%
\tempurl}


\bibitem[\protect\citeauthoryear{Li, Kadav, Durdanovic, Samet, and Graf}{Li
  et~al\mbox{.}}{2017b}]%
        {pruning:2017}
\bibfield{author}{\bibinfo{person}{Hao Li}, \bibinfo{person}{Asim Kadav},
  \bibinfo{person}{Igor Durdanovic}, \bibinfo{person}{Hanan Samet}, {and}
  \bibinfo{person}{Hans~Peter Graf}.} \bibinfo{year}{2017}\natexlab{b}.
\newblock \showarticletitle{Pruning Filters for Efficient ConvNets}. In
  \bibinfo{booktitle}{\emph{5th International Conference on Learning
  Representations, {ICLR} 2017, Toulon, France, April 24-26, 2017, Conference
  Track Proceedings}}. \bibinfo{publisher}{OpenReview.net}.
\newblock
\urldef\tempurl%
\url{https://openreview.net/forum?id=rJqFGTslg}
\showURL{%
\tempurl}


\bibitem[\protect\citeauthoryear{Li, Jamieson, DeSalvo, Rostamizadeh, and
  Talwalkar}{Li et~al\mbox{.}}{2017a}]%
        {HB:2017}
\bibfield{author}{\bibinfo{person}{Lisha Li}, \bibinfo{person}{Kevin Jamieson},
  \bibinfo{person}{Giulia DeSalvo}, \bibinfo{person}{Afshin Rostamizadeh},
  {and} \bibinfo{person}{Ameet Talwalkar}.} \bibinfo{year}{2017}\natexlab{a}.
\newblock \showarticletitle{Hyperband: A Novel Bandit-Based Approach to
  Hyperparameter Optimization}.
\newblock  \bibinfo{volume}{18}, \bibinfo{number}{1} (\bibinfo{date}{1}
  \bibinfo{year}{2017}), \bibinfo{pages}{6765–6816}.
\newblock
\showISSN{1532-4435}


\bibitem[\protect\citeauthoryear{Li, Jamieson, Rostamizadeh, Gonina, Ben-tzur,
  Hardt, Recht, and Talwalkar}{Li et~al\mbox{.}}{2020}]%
        {AHB:2020}
\bibfield{author}{\bibinfo{person}{Liam Li}, \bibinfo{person}{Kevin Jamieson},
  \bibinfo{person}{Afshin Rostamizadeh}, \bibinfo{person}{Ekaterina Gonina},
  \bibinfo{person}{Jonathan Ben-tzur}, \bibinfo{person}{Moritz Hardt},
  \bibinfo{person}{Benjamin Recht}, {and} \bibinfo{person}{Ameet Talwalkar}.}
  \bibinfo{year}{2020}\natexlab{}.
\newblock \showarticletitle{A System for Massively Parallel Hyperparameter
  Tuning}. In \bibinfo{booktitle}{\emph{Proceedings of Machine Learning and
  Systems}}, \bibfield{editor}{\bibinfo{person}{I.~Dhillon},
  \bibinfo{person}{D.~Papailiopoulos}, {and} \bibinfo{person}{V.~Sze}} (Eds.),
  Vol.~\bibinfo{volume}{2}. \bibinfo{pages}{230--246}.
\newblock
\urldef\tempurl%
\url{https://proceedings.mlsys.org/paper/2020/file/f4b9ec30ad9f68f89b29639786cb62ef-Paper.pdf}
\showURL{%
\tempurl}


\bibitem[\protect\citeauthoryear{Li, Khodak, Balcan, and Talwalkar}{Li
  et~al\mbox{.}}{2021a}]%
        {rsps:2020}
\bibfield{author}{\bibinfo{person}{Liam Li}, \bibinfo{person}{Mikhail Khodak},
  \bibinfo{person}{Nina Balcan}, {and} \bibinfo{person}{Ameet Talwalkar}.}
  \bibinfo{year}{2021}\natexlab{a}.
\newblock \showarticletitle{Geometry-Aware Gradient Algorithms for Neural
  Architecture Search}. In \bibinfo{booktitle}{\emph{International Conference
  on Learning Representations}}.
\newblock
\urldef\tempurl%
\url{https://openreview.net/forum?id=MuSYkd1hxRP}
\showURL{%
\tempurl}


\bibitem[\protect\citeauthoryear{Li and Talwalkar}{Li and Talwalkar}{2020}]%
        {random:2020}
\bibfield{author}{\bibinfo{person}{Liam Li} {and} \bibinfo{person}{Ameet
  Talwalkar}.} \bibinfo{year}{2020}\natexlab{}.
\newblock \showarticletitle{Random Search and Reproducibility for Neural
  Architecture Search}. In \bibinfo{booktitle}{\emph{Proceedings of The 35th
  Uncertainty in Artificial Intelligence Conference}}
  \emph{(\bibinfo{series}{Proceedings of Machine Learning Research},
  Vol.~\bibinfo{volume}{115})}, \bibfield{editor}{\bibinfo{person}{Ryan~P.
  Adams} {and} \bibinfo{person}{Vibhav Gogate}} (Eds.).
  \bibinfo{publisher}{PMLR}, \bibinfo{pages}{367--377}.
\newblock
\urldef\tempurl%
\url{http://proceedings.mlr.press/v115/li20c.html}
\showURL{%
\tempurl}


\bibitem[\protect\citeauthoryear{Liao and Moody}{Liao and Moody}{1999}]%
        {heteroinput:1999}
\bibfield{author}{\bibinfo{person}{Yuansong Liao} {and} \bibinfo{person}{John
  Moody}.} \bibinfo{year}{1999}\natexlab{}.
\newblock \showarticletitle{Constructing Heterogeneous Committees Using Input
  Feature Grouping: Application to Economic Forecasting}.
\newblock  (\bibinfo{year}{1999}), \bibinfo{pages}{921–927}.
\newblock


\bibitem[\protect\citeauthoryear{Liaw}{Liaw}{2019}]%
        {tune:2018}
\bibfield{author}{\bibinfo{person}{Richard Liaw}.}
  \bibinfo{year}{2019}\natexlab{}.
\newblock \showarticletitle{{A Guide to Modern Hyperparameters Turning
  Algorithms}}. In \bibinfo{booktitle}{\emph{PyData Los Angeles}}.
\newblock


\bibitem[\protect\citeauthoryear{Liu, Simonyan, and Yang}{Liu
  et~al\mbox{.}}{2019}]%
        {darts:2018}
\bibfield{author}{\bibinfo{person}{Hanxiao Liu}, \bibinfo{person}{Karen
  Simonyan}, {and} \bibinfo{person}{Yiming Yang}.}
  \bibinfo{year}{2019}\natexlab{}.
\newblock \showarticletitle{{DARTS}: Differentiable Architecture Search}. In
  \bibinfo{booktitle}{\emph{International Conference on Learning
  Representations}}.
\newblock
\urldef\tempurl%
\url{https://openreview.net/forum?id=S1eYHoC5FX}
\showURL{%
\tempurl}


\bibitem[\protect\citeauthoryear{Moritz, Nishihara, Wang, Tumanov, Liaw, Liang,
  Elibol, Yang, Paul, Jordan, and Stoica}{Moritz et~al\mbox{.}}{2018}]%
        {ray}
\bibfield{author}{\bibinfo{person}{Philipp Moritz}, \bibinfo{person}{Robert
  Nishihara}, \bibinfo{person}{Stephanie Wang}, \bibinfo{person}{Alexey
  Tumanov}, \bibinfo{person}{Richard Liaw}, \bibinfo{person}{Eric Liang},
  \bibinfo{person}{Melih Elibol}, \bibinfo{person}{Zongheng Yang},
  \bibinfo{person}{William Paul}, \bibinfo{person}{Michael~I. Jordan}, {and}
  \bibinfo{person}{Ion Stoica}.} \bibinfo{year}{2018}\natexlab{}.
\newblock \showarticletitle{Ray: A Distributed Framework for Emerging AI
  Applications}. In \bibinfo{booktitle}{\emph{Proceedings of the 13th USENIX
  Conference on Operating Systems Design and Implementation}} (Carlsbad, CA,
  USA) \emph{(\bibinfo{series}{OSDI'18})}. \bibinfo{publisher}{USENIX
  Association}, \bibinfo{address}{USA}, \bibinfo{pages}{561–577}.
\newblock
\showISBNx{9781931971478}


\bibitem[\protect\citeauthoryear{Oktay, Ball{\'{e}}, Singh, and
  Shrivastava}{Oktay et~al\mbox{.}}{2020}]%
        {compressentropy:2020}
\bibfield{author}{\bibinfo{person}{Deniz Oktay}, \bibinfo{person}{Johannes
  Ball{\'{e}}}, \bibinfo{person}{Saurabh Singh}, {and} \bibinfo{person}{Abhinav
  Shrivastava}.} \bibinfo{year}{2020}\natexlab{}.
\newblock \showarticletitle{Scalable Model Compression by Entropy Penalized
  Reparameterization}. In \bibinfo{booktitle}{\emph{8th International
  Conference on Learning Representations, {ICLR} 2020, Addis Ababa, Ethiopia,
  April 26-30, 2020}}. \bibinfo{publisher}{OpenReview.net}.
\newblock
\urldef\tempurl%
\url{https://openreview.net/forum?id=HkgxW0EYDS}
\showURL{%
\tempurl}


\bibitem[\protect\citeauthoryear{Olston, Li, Harmsen, Soyke, Gorovoy, Lao,
  Fiedel, Ramesh, and Rajashekhar}{Olston et~al\mbox{.}}{2017}]%
        {tfserv}
\bibfield{author}{\bibinfo{person}{Christopher Olston},
  \bibinfo{person}{Fangwei Li}, \bibinfo{person}{Jeremiah Harmsen},
  \bibinfo{person}{Jordan Soyke}, \bibinfo{person}{Kiril Gorovoy},
  \bibinfo{person}{Li Lao}, \bibinfo{person}{Noah Fiedel},
  \bibinfo{person}{Sukriti Ramesh}, {and} \bibinfo{person}{Vinu Rajashekhar}.}
  \bibinfo{year}{2017}\natexlab{}.
\newblock \showarticletitle{TensorFlow-Serving: Flexible, High-Performance ML
  Serving}.
\newblock
\newblock
\shownote{Workshop on ML Systems at NIPS 2017.}


\bibitem[\protect\citeauthoryear{Pathak, Cai, and Rajasekaran}{Pathak
  et~al\mbox{.}}{2018}]%
        {enstim:2018}
\bibfield{author}{\bibinfo{person}{Sudipta Pathak}, \bibinfo{person}{Xingyu
  Cai}, {and} \bibinfo{person}{Sanguthevar Rajasekaran}.}
  \bibinfo{year}{2018}\natexlab{}.
\newblock \showarticletitle{Ensemble Deep TimeNet: An Ensemble Learning
  Approach with Deep Neural Networks for Time Series}. In
  \bibinfo{booktitle}{\emph{2018 IEEE 8th International Conference on
  Computational Advances in Bio and Medical Sciences (ICCABS)}}.
  \bibinfo{pages}{1--1}.
\newblock
\urldef\tempurl%
\url{https://doi.org/10.1109/ICCABS.2018.8541985}
\showDOI{\tempurl}


\bibitem[\protect\citeauthoryear{Pham, Guan, Zoph, Le, and Dean}{Pham
  et~al\mbox{.}}{2018}]%
        {ENAS:2018}
\bibfield{author}{\bibinfo{person}{Hieu Pham}, \bibinfo{person}{Melody Guan},
  \bibinfo{person}{Barret Zoph}, \bibinfo{person}{Quoc Le}, {and}
  \bibinfo{person}{Jeff Dean}.} \bibinfo{year}{2018}\natexlab{}.
\newblock \showarticletitle{Efficient Neural Architecture Search via Parameters
  Sharing}. In \bibinfo{booktitle}{\emph{Proceedings of the 35th International
  Conference on Machine Learning}} \emph{(\bibinfo{series}{Proceedings of
  Machine Learning Research}, Vol.~\bibinfo{volume}{80})},
  \bibfield{editor}{\bibinfo{person}{Jennifer Dy} {and}
  \bibinfo{person}{Andreas Krause}} (Eds.). \bibinfo{publisher}{PMLR},
  \bibinfo{pages}{4095--4104}.
\newblock
\urldef\tempurl%
\url{https://proceedings.mlr.press/v80/pham18a.html}
\showURL{%
\tempurl}


\bibitem[\protect\citeauthoryear{Prechelt}{Prechelt}{1998}]%
        {early:1998}
\bibfield{author}{\bibinfo{person}{Lutz Prechelt}.}
  \bibinfo{year}{1998}\natexlab{}.
\newblock \showarticletitle{Early Stopping-But When?}. In
  \bibinfo{booktitle}{\emph{Neural Networks: Tricks of the Trade, This Book is
  an Outgrowth of a 1996 NIPS Workshop}}. \bibinfo{publisher}{Springer-Verlag},
  \bibinfo{address}{Berlin, Heidelberg}, \bibinfo{pages}{55–69}.
\newblock
\showISBNx{3540653112}


\bibitem[\protect\citeauthoryear{Real, Moore, Selle, Saxena, Suematsu, Tan, Le,
  and Kurakin}{Real et~al\mbox{.}}{2017}]%
        {evolution:2017}
\bibfield{author}{\bibinfo{person}{Esteban Real}, \bibinfo{person}{Sherry
  Moore}, \bibinfo{person}{Andrew Selle}, \bibinfo{person}{Saurabh Saxena},
  \bibinfo{person}{Yutaka~Leon Suematsu}, \bibinfo{person}{Jie Tan},
  \bibinfo{person}{Quoc~V. Le}, {and} \bibinfo{person}{Alexey Kurakin}.}
  \bibinfo{year}{2017}\natexlab{}.
\newblock \showarticletitle{Large-Scale Evolution of Image Classifiers}. In
  \bibinfo{booktitle}{\emph{Proceedings of the 34th International Conference on
  Machine Learning - Volume 70}} (Sydney, NSW, Australia)
  \emph{(\bibinfo{series}{ICML'17})}. \bibinfo{publisher}{JMLR.org},
  \bibinfo{pages}{2902–2911}.
\newblock


\bibitem[\protect\citeauthoryear{Snoek, Larochelle, and Adams}{Snoek
  et~al\mbox{.}}{2012}]%
        {snoek:2012}
\bibfield{author}{\bibinfo{person}{Jasper Snoek}, \bibinfo{person}{Hugo
  Larochelle}, {and} \bibinfo{person}{Ryan~P Adams}.}
  \bibinfo{year}{2012}\natexlab{}.
\newblock \showarticletitle{Practical Bayesian Optimization of Machine Learning
  Algorithms}. In \bibinfo{booktitle}{\emph{Advances in Neural Information
  Processing Systems}}, \bibfield{editor}{\bibinfo{person}{F.~Pereira},
  \bibinfo{person}{C.~J.~C. Burges}, \bibinfo{person}{L.~Bottou}, {and}
  \bibinfo{person}{K.~Q. Weinberger}} (Eds.), Vol.~\bibinfo{volume}{25}.
  \bibinfo{publisher}{Curran Associates, Inc.}, \bibinfo{pages}{2951--2959}.
\newblock
\urldef\tempurl%
\url{https://proceedings.neurips.cc/paper/2012/file/05311655a15b75fab86956663e1819cd-Paper.pdf}
\showURL{%
\tempurl}


\bibitem[\protect\citeauthoryear{Sollich and Krogh}{Sollich and Krogh}{1995}]%
        {useoverfit:1995}
\bibfield{author}{\bibinfo{person}{Peter Sollich} {and} \bibinfo{person}{Anders
  Krogh}.} \bibinfo{year}{1995}\natexlab{}.
\newblock \showarticletitle{Learning with ensembles: How overfitting can be
  useful.}
\newblock   \bibinfo{volume}{8} (\bibinfo{date}{01} \bibinfo{year}{1995}),
  \bibinfo{pages}{190--196}.
\newblock


\bibitem[\protect\citeauthoryear{Sun, Liu, Mengxue, Zhongxin, Jia, and
  Xiaowen}{Sun et~al\mbox{.}}{2020}]%
        {enstex:2020}
\bibfield{author}{\bibinfo{person}{Gang Sun}, \bibinfo{person}{Jianqiao Liu},
  \bibinfo{person}{Wei Mengxue}, \bibinfo{person}{Wang Zhongxin},
  \bibinfo{person}{Zhao Jia}, {and} \bibinfo{person}{Guan Xiaowen}.}
  \bibinfo{year}{2020}\natexlab{}.
\newblock \showarticletitle{An Ensemble Classification Algorithm for Imbalanced
  Text Data Streams}. In \bibinfo{booktitle}{\emph{2020 IEEE International
  Conference on Artificial Intelligence and Computer Applications (ICAICA)}}.
  \bibinfo{pages}{1073--1076}.
\newblock
\urldef\tempurl%
\url{https://doi.org/10.1109/ICAICA50127.2020.9182576}
\showDOI{\tempurl}


\bibitem[\protect\citeauthoryear{Sun and Ge}{Sun and Ge}{2021}]%
        {enssem:2021}
\bibfield{author}{\bibinfo{person}{Qingqiang Sun} {and}
  \bibinfo{person}{Zhiqiang Ge}.} \bibinfo{year}{2021}\natexlab{}.
\newblock \showarticletitle{Deep Learning for Industrial KPI Prediction: When
  Ensemble Learning Meets Semi-Supervised Data}.
\newblock \bibinfo{journal}{\emph{IEEE Transactions on Industrial Informatics}}
  \bibinfo{volume}{17}, \bibinfo{number}{1} (\bibinfo{year}{2021}),
  \bibinfo{pages}{260--269}.
\newblock
\urldef\tempurl%
\url{https://doi.org/10.1109/TII.2020.2969709}
\showDOI{\tempurl}


\bibitem[\protect\citeauthoryear{Tan and Le}{Tan and Le}{2019}]%
        {eff:2019}
\bibfield{author}{\bibinfo{person}{Mingxing Tan} {and} \bibinfo{person}{Quoc
  Le}.} \bibinfo{year}{2019}\natexlab{}.
\newblock \showarticletitle{{E}fficient{N}et: Rethinking Model Scaling for
  Convolutional Neural Networks}. In \bibinfo{booktitle}{\emph{Proceedings of
  the 36th International Conference on Machine Learning}}
  \emph{(\bibinfo{series}{Proceedings of Machine Learning Research},
  Vol.~\bibinfo{volume}{97})}, \bibfield{editor}{\bibinfo{person}{Kamalika
  Chaudhuri} {and} \bibinfo{person}{Ruslan Salakhutdinov}} (Eds.).
  \bibinfo{publisher}{PMLR}, \bibinfo{pages}{6105--6114}.
\newblock
\urldef\tempurl%
\url{https://proceedings.mlr.press/v97/tan19a.html}
\showURL{%
\tempurl}


\bibitem[\protect\citeauthoryear{Thornton, Hutter, Hoos, and
  Leyton-Brown}{Thornton et~al\mbox{.}}{2012}]%
        {autoweka}
\bibfield{author}{\bibinfo{person}{Chris Thornton}, \bibinfo{person}{Frank
  Hutter}, \bibinfo{person}{Holger Hoos}, {and} \bibinfo{person}{Kevin
  Leyton-Brown}.} \bibinfo{year}{2012}\natexlab{}.
\newblock \showarticletitle{Auto-WEKA: Combined Selection and Hyperparameter
  Optimization of Classification Algorithms}.
\newblock \bibinfo{journal}{\emph{KDD}} (\bibinfo{date}{08}
  \bibinfo{year}{2012}).
\newblock
\urldef\tempurl%
\url{https://doi.org/10.1145/2487575.2487629}
\showDOI{\tempurl}


\bibitem[\protect\citeauthoryear{Tsoumakas, Partalas, and Vlahavas}{Tsoumakas
  et~al\mbox{.}}{2008}]%
        {tsoumakas:2008}
\bibfield{author}{\bibinfo{person}{Grigorios Tsoumakas},
  \bibinfo{person}{Ioannis Partalas}, {and} \bibinfo{person}{I. Vlahavas}.}
  \bibinfo{year}{2008}\natexlab{}.
\newblock \showarticletitle{A Taxonomy and Short Review of Ensemble Selection}.
\newblock \bibinfo{journal}{\emph{ECAI 2008, Workshop on Supervised and
  Unsupervised Ensemble Methods and Their Applications}} (\bibinfo{date}{01}
  \bibinfo{year}{2008}).
\newblock


\bibitem[\protect\citeauthoryear{Wolpert and Macready}{Wolpert and
  Macready}{1997}]%
        {NFL:1997}
\bibfield{author}{\bibinfo{person}{D.~H. Wolpert} {and} \bibinfo{person}{W.~G.
  Macready}.} \bibinfo{year}{1997}\natexlab{}.
\newblock \showarticletitle{No Free Lunch Theorems for Optimization}.
\newblock \bibinfo{journal}{\emph{Trans. Evol. Comp}} \bibinfo{volume}{1},
  \bibinfo{number}{1} (\bibinfo{date}{4} \bibinfo{year}{1997}),
  \bibinfo{pages}{67–82}.
\newblock
\showISSN{1089-778X}
\urldef\tempurl%
\url{https://doi.org/10.1109/4235.585893}
\showDOI{\tempurl}


\bibitem[\protect\citeauthoryear{Xie, Girshick, Dollár, Tu, and He}{Xie
  et~al\mbox{.}}{2017}]%
        {resnext:2016}
\bibfield{author}{\bibinfo{person}{Saining Xie}, \bibinfo{person}{Ross
  Girshick}, \bibinfo{person}{Piotr Dollár}, \bibinfo{person}{Zhuowen Tu},
  {and} \bibinfo{person}{Kaiming He}.} \bibinfo{year}{2017}\natexlab{}.
\newblock \showarticletitle{Aggregated Residual Transformations for Deep Neural
  Networks}. In \bibinfo{booktitle}{\emph{2017 IEEE Conference on Computer
  Vision and Pattern Recognition (CVPR)}}. \bibinfo{pages}{5987--5995}.
\newblock
\urldef\tempurl%
\url{https://doi.org/10.1109/CVPR.2017.634}
\showDOI{\tempurl}


\bibitem[\protect\citeauthoryear{Xu}{Xu}{2020}]%
        {tritonserv}
\bibfield{author}{\bibinfo{person}{Tianhao Xu}.}
  \bibinfo{year}{2020}\natexlab{}.
\newblock \showarticletitle{“Deep into Triton Inference Server: BERT
  Practical Deployment on NVIDIA GPU”}.
\newblock
\newblock
\shownote{GPU Technology Conference.}


\bibitem[\protect\citeauthoryear{Yanan, Bing, Mengjie, G., and Jiancheng}{Yanan
  et~al\mbox{.}}{2020}]%
        {evolution:2020}
\bibfield{author}{\bibinfo{person}{Sun Yanan}, \bibinfo{person}{Xue Bing},
  \bibinfo{person}{Zhang Mengjie}, \bibinfo{person}{Yen~Gary G.}, {and}
  \bibinfo{person}{Lv Jiancheng}.} \bibinfo{year}{2020}\natexlab{}.
\newblock \showarticletitle{Automatically Designing CNN Architectures Using the
  Genetic Algorithm for Image Classification}.
\newblock \bibinfo{journal}{\emph{IEEE Transactions on Cybernetics}}
  \bibinfo{volume}{50}, \bibinfo{number}{9} (\bibinfo{year}{2020}),
  \bibinfo{pages}{3840--3854}.
\newblock
\urldef\tempurl%
\url{https://doi.org/10.1109/TCYB.2020.2983860}
\showDOI{\tempurl}


\bibitem[\protect\citeauthoryear{Zagoruyko and Komodakis}{Zagoruyko and
  Komodakis}{2016}]%
        {widnet:2016}
\bibfield{author}{\bibinfo{person}{Sergey Zagoruyko} {and}
  \bibinfo{person}{Nikos Komodakis}.} \bibinfo{year}{2016}\natexlab{}.
\newblock \showarticletitle{Wide Residual Networks}. In
  \bibinfo{booktitle}{\emph{Proceedings of the British Machine Vision
  Conference 2016 (BMVC 2016)}}, \bibfield{editor}{\bibinfo{person}{Richard~C.
  Wilson}, \bibinfo{person}{Edwin~R. Hancock}, {and} \bibinfo{person}{William
  A.~P. Smith}} (Eds.). \bibinfo{publisher}{BMVA Press}.
\newblock
\urldef\tempurl%
\url{http://www.bmva.org/bmvc/2016/papers/paper087/index.html}
\showURL{%
\tempurl}


\bibitem[\protect\citeauthoryear{Zhang, Zou, He, and Sun}{Zhang
  et~al\mbox{.}}{2016}]%
        {VGG:2014}
\bibfield{author}{\bibinfo{person}{Xiangyu Zhang}, \bibinfo{person}{Jianhua
  Zou}, \bibinfo{person}{Kaiming He}, {and} \bibinfo{person}{Jian Sun}.}
  \bibinfo{year}{2016}\natexlab{}.
\newblock \showarticletitle{Accelerating Very Deep Convolutional Networks for
  Classification and Detection}.
\newblock \bibinfo{journal}{\emph{IEEE Trans. Pattern Anal. Mach. Intell.}}
  \bibinfo{volume}{38}, \bibinfo{number}{10} (\bibinfo{date}{Oct.}
  \bibinfo{year}{2016}), \bibinfo{pages}{1943–1955}.
\newblock
\showISSN{0162-8828}
\urldef\tempurl%
\url{https://doi.org/10.1109/TPAMI.2015.2502579}
\showDOI{\tempurl}


\end{thebibliography}
%\printbibliography
%\bibliography{bib_automl}

\clearpage
\appendix

\section{Experimental data set and hyperparameter settings}
\label{sec:appdesign}

This section describes machine learning experiments for reproducibility purposes.

\subsection{The two datasets used}
\label{sec:db}

\textbf{The CIFAR100 dataset.} CIFAR100~\cite{cifardb:2009} consists to 60,000 32x32 RGB images in 100 classes. For each class, there are 580 training images, 20 validation images and 100 testing images.

\textbf{The Microfossils dataset.} Microfossils are extremely useful in age dating, correlation, and paleo-environmental reconstruction to refine our knowledge of geology. Microfossil species are identified and counted on large microscope images and thanks to their frequencies we can compute the date of sedimentary rocks. 

% big data context
To do reliable statistics, a big number of objects needs to be identified. That is why we need deep learning to automate this work. Today, thousands of fields of view (microscopy imagery) need to be shot for 1 rock sample. In each field of view, there are hundreds of objects to identify. Among these objects, there are non-fossils (crystals, rock grains, etc...) and others are fossils that we are looking for to study rocks.

Our dataset contains 91 classes of 224x224 RGB images (after homemade preprocessing). Microfossils are calcareous objects took with polarized light microscopy. The classes are unbalanced, we have from 50 images to 2500 images by class, with a total of 32K images in all the datasets. The train/validation/test split is as following: 72\% 8\% 20\%. The F1 score was used and labeled as 'accuracy' on all benchmarks.

%\subsection{Hyperparameter optimization settings}
%Hyperband strategies do not require
% To train them after a fixed number of times, hyperband strategies (HB and BOHB) are used with x3 more samplings. 
%Several tricks: halt after first epochs if accuracy is not better. Moreover, we implemented more standard early stopping consisting to stop when the plateau 
%For a fair comparison between strategies presented they all benefit of t
%Some strategies are more efficient than others in terms of computing power. For a fair comparison between them and the same running duration, they are set with the same

\subsection{Hyperparameter configuration space}
\label{sec:hpospace}

Table~\ref {tab:configspace} shows all hyperparameters properties in this workflow. We use ResNet-based architectures due to their simplicity to yield promising and robust models on many datasets. We explore different residual block versions: "V1", "V2" \cite{resnet:2015} and "next" \cite{resnext:2016}.  Regarding the optimization method, we use Adam optimizer \cite{adam:2014} due to its well-known performance and its lower learning rate tuning requirement.

\begin{table}
\centering
\small
\begin{tabularx}{\linewidth}{lXcccX}
\toprule
Category                                            & Name                                               & Type        & Range                                  \\ \midrule

\multirow{3}{*}{Optimization}                & Learning rate                                      & Continuous  & [0.001; 0.01]   \\
                                             \cmidrule{2-4}
                                             & Batch size                                         & Discrete    & [8; 48]                  &                       \\
                                             \cmidrule{2-4}
                                             & L2 regularization factor                           & Continuous  & [0; 0.1] &                       \\ 
\midrule
\multirow{8}{*}{\shortstack[c]{NN \\ architecture}} & Convolution type                                   & Categorical & \{v1,v2,next\}         \\
                                             \cmidrule{2-4}
                                             & Activation function                                & Categorical & \{tanh,relu,elu\}    &                       \\
                                             \cmidrule{2-4}
                                             & Number of filters in the first convolutional layer   & Discrete    & [32; 128]               &                       \\
                                             \cmidrule{2-4}
                                             & Multiplier of filters in the 4 blocks      & Discrete    & \shortstack{[32; 128] }         &                       \\
                                             \cmidrule{2-4}
                                             & Number of convolutional block in the first stage  & Discrete    & [1;11]             &                       \\
                                             \cmidrule{2-4}
                                             & Number of convolutional block in the second stage & Discrete    & [1;11]               &                      \\
                                             \cmidrule{2-4}
                                             & Number of convolutional block in the third stage  & Discrete    & [1;11]                &                      \\
                                             \cmidrule{2-4}
                                             & Number of convolutional block in the fourth stage & Discrete    &  [1;11]               &                       \\ \midrule
\multirow{5}{*}{\shortstack{Data \\ augmentation}}           & Max zoom                                           & Continuous  & [0; 0.6]  \\
                                             \cmidrule{2-4}
                                             & Max translation                                    & Continuous  & [0; 0.6]                  &                       \\
                                             \cmidrule{2-4}
                                             & Max shearing                                       & Continuous  & [0; 0.3]                  &                       \\
                                             \cmidrule{2-4}
                                             & Max channel shifting                               & Continuous  & [0; 0.3]                  &                       \\
                                             \cmidrule{2-4}
                                             & Max rotation measured in degrees                   & Discrete    & [0; 90]&                      \\
\bottomrule
\end{tabularx}
\caption{The hyperparameter space experimented based on the ResNet neural architecture framework}
\label{tab:configspace}
\end{table}

The simplicity of the ResNet architecture makes this work easy to test by a scientist on its image dataset \cite{widnet:2016}. Exploring other convolutional block type like VGG \cite{VGG:2014}, Xception \cite{Xception:2016}, DenseNet \cite{densenet:2016} and EfficientNet \cite{eff:2019} are potential improvement which could increase the degree of liberty in DNNs construction and improve again the accuracy of ensembles. 

% adaptation resnet50
The CIFAR100 dataset contains 32x32 images while usually ResNet is adapted to be used on ImageNet (224x244 images). Those different resolutions need some adaptation. Therefore, on the CIFAR100 case, the first convolutional network is replaced from the 7x7 kernel size with a stride of 2, to a 3x3 kernel size with a stride of 1. With equivalent settings, our CNN framework produces nearly the same number of weights between the CIFAR100 and microfossils dataset, but the time complexity is factor 11 different because of different signal resolutions flowing through the layers.

\end{document}